%% file: arxiv_submit_iclr2027_conference.tex
\documentclass{article} 
\usepackage{arxiv_submit_iclr2027_conference,times}

\input{math_commands.tex}

\usepackage{hyperref}
\hypersetup{colorlinks=true, citecolor=[rgb]{0.1,0.3,0.6}, linkcolor=[rgb]{0.6,0.1,0.1}, urlcolor=[rgb]{0.1,0.3,0.6}}
\usepackage{url}

\usepackage{amsmath,amssymb}
\usepackage{booktabs}
\usepackage{multirow}
\usepackage{xcolor}
\usepackage{colortbl}
\usepackage{longtable}
\renewcommand{\eqref}[1]{(\ref{#1})}

\usepackage{graphicx}
\usepackage{wrapfig}
\usepackage{placeins}
\usepackage{pifont}
\newcommand{\cmark}{\textcolor[HTML]{2E7D32}{\ding{51}}}
\newcommand{\xmark}{\textcolor[HTML]{C62828}{\ding{55}}}

\title{Custom Forcing: Training-Free Subject Customization for Autoregressive Video Generation}

\author{
Yunseung Ok$^{*1}$ \quad
Hyunsoo Kim$^{*2}$ \quad
Minseo Kim$^{1}$ \quad
Suhyun Kim$^{\dagger 1}$ \\
$^{1}$Kyung Hee University \qquad
$^{2}$The University of Texas at Austin \\
$^{*}$Equal contribution \qquad
$^{\dagger}$Corresponding author
}

\iclrfinalcopy
\begin{document}

\maketitle

\begin{center}
    \includegraphics[width=\linewidth]{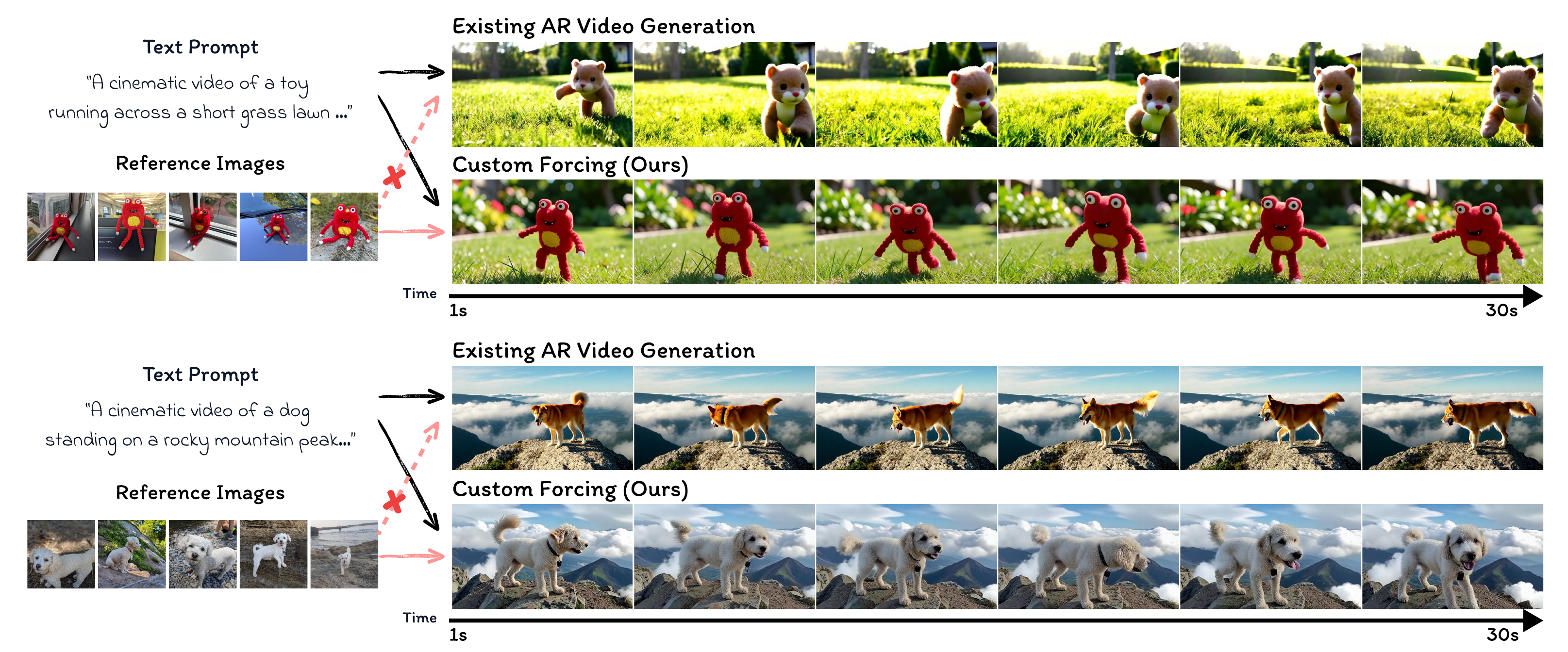}
\end{center}

\vspace{-1em}

\refstepcounter{figure}
\label{fig:overview_figure}

\makeatletter
\@makecaption{\fnum@figure}{
    \textbf{Comparison with existing autoregressive video generation.} For each text prompt, an existing autoregressive video model (top) generates a generic instance of the target category, whereas Custom Forcing (bottom) also conditions on a few reference images and preserves the identity of the specific subject throughout the video without any training. Video demos are available on our
\href{https://gustn9609.github.io/custom-forcing/}{project page}.
}
\makeatother

\vspace{0.5em}

\begin{abstract}
Autoregressive video models can generate minute-long videos in real time, but they produce generic subjects from text rather than specific subjects from user-provided images. Existing customization methods either require costly per-subject optimization or use pretrained conditioning networks that jointly process all video frames with bidirectional attention. Neither approach is designed for causal streaming. We present \textbf{Custom Forcing}, a training-free method that stores reference-based anchor frames in the persistent KV cache of a frozen autoregressive video model. However, fixed anchors face two limitations: simple conditioning allows identity to drift, and the text prompt continues to favor a generic subject. To address these problems, \emph{drift-adaptive value amplification} (DVA) scales reference influence with the degree of identity drift, while \emph{anchor contrast guidance} (ACG) steers generation away from the generic class prior. Over two-minute rollouts, fixed anchors fall from $0.58$ to $0.42$ in DINO-I, while Custom Forcing keeps it between $0.58$ and $0.62$ without reducing motion. Custom Forcing also achieves higher subject similarity than bidirectional customization methods and better preserves identity over $30$s than causal image-to-video and reference-to-video models, while generating each frame $9.5$--$28.5\times$ faster than these long-video baselines. 
\end{abstract}


\section{Introduction}
\label{sec:intro}

Video generation models can now turn a text prompt or an input image into a high-quality video, and they are increasingly used for world models, robotics, and content creation~\citep{videoworldsimulators2024,wan,nextforcing,arforcing,actionparty}. Bidirectional video generation models~\citep{videoworldsimulators2024,wan} denoise all frames jointly at every sampling step, so every step processes the entire video and generation is slow. Autoregressive video models instead generate frames or chunks in temporal order and condition each step on previous outputs~\citep{diffusionforcing,magi1}. With distillation or adversarial post-training, they need only a few denoising steps per chunk and can stream video in real time~\citep{causvid,selfforcing,apt2}, and rolling contexts bound their memory as the video grows~\citep{rollingforcing,longlive,selfforcingpp,framepack}. These properties support real-time and long-horizon applications such as interactive video generation and world models~\citep{longlive,nextforcing,arforcing,actionparty}. In such applications, users often want a specific subject, such as their own pet or toy, to appear throughout the video. Given only a text prompt, however, a conventional autoregressive video model generates a generic instance of the category.

Image customization methods either optimize the model for each subject~\citep{textualinversion,dreambooth,customdiffusion} or use pretrained subject encoders or adapters that generalize to unseen subjects~\citep{elite,ipadapter,photomaker,instantid}. Video customization follows the same two approaches~\citep{magicme,dreamvideo,customcrafter,videobooth,consisid,standin}, but optimizing a video model for each subject is costly, and training with still images makes it difficult to preserve motion~\citep{stillmoving,customcrafter}. These methods also denoise the frames of a bounded video jointly, which a causal generator cannot do: its finalized frames cannot be revised, and the reference must remain available throughout an open-ended rollout. A simpler alternative gives an image of the subject to an image-to-video model as its first frame. As Sec.~\ref{sec:exp:main} shows, however, causal image-to-video models lose the subject within $30$\,s, and a bidirectional one cannot generate long videos.

\begin{wrapfigure}{r}{0.36\linewidth} \vspace{-14pt} \centering \includegraphics[width=\linewidth]{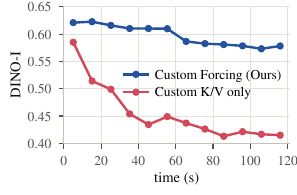} \vspace{-18pt} \caption{\textbf{DINO-I over two minutes}, $n=100$. Custom K/V only uses fixed anchor frames without DVA or ACG.} \label{fig:intro} \vspace{-12pt} \end{wrapfigure} In streaming generation, once a chunk is finalized, appearance errors propagate through subsequent chunks and cause progressive divergence from the external subject reference. We call this \emph{identity drift}. Existing long-video methods maintain consistency with generated history using attention sinks~\citep{rollingforcing,rollingsink,deepforcing,dysink,relaxforcing}, persistent memory~\citep{anchorforcing,memrope,infinityrope,adastate}, or longer contexts~\citep{selfforcingpp,longlive,contextforcing,packforcing}. However, consistency with generated history does not ensure fidelity to the user-provided subject. We store the user's reference images and a scene-specific \emph{custom anchor} as persistent anchor frames in the KV cache, making them available to every generated chunk without fine-tuning or an additional conditioning network. These anchors establish the subject initially but cannot prevent identity drift: as shown by the red curve in Fig.~\ref{fig:intro}, identity similarity steadily declines over a two-minute rollout. Weak anchor influence loses subject-specific details, and strong influence suppresses motion. The text prompt also biases generation toward a generic instance of the subject category.

We present \textbf{Custom Forcing}, which addresses these limitations by modifying only the backbone's self-attention. \emph{Drift-adaptive value amplification} (DVA) scales reference influence with the measured drift of each finalized chunk. \emph{Anchor contrast guidance} (ACG) contrasts attention outputs computed with and without the anchor frames and extrapolates along their difference, counteracting the generic class prior induced by the prompt.

Over two-minute rollouts, Custom Forcing maintains DINO-I between $0.58$ and $0.62$ (blue curve), while fixed anchors decline from $0.58$ to $0.42$ (red curve, Fig.~\ref{fig:intro}). The improvement grows over time, reaching $+0.163$ DINO-I in the final window without reducing motion. Custom Forcing also better preserves subject identity than causal image-to-video baselines over $30$\,s, with only $11$--$13\%$ overhead over fixed anchors.

Our contributions are as follows:
\begin{itemize}
    \item We present Custom Forcing, a training-free method that customizes a frozen autoregressive video model by keeping the reference images and a custom anchor in its KV cache as anchor frames, without fine-tuning or an additional conditioning network.
    \item We introduce DVA and ACG, two self-attention operations that keep the subject over long rollouts. DVA scales the reference influence with the measured drift of finalized chunks, and ACG extrapolates away from the attention output computed without the anchors to counter the generic class prior.
    \item We show that Custom Forcing keeps DINO-I between $0.58$ and $0.62$ over two-minute videos, where fixed anchors fall to $0.42$, with comparable motion. It exceeds bidirectional customization methods and causal image-to-video models, and matches a 14B reference-to-video model with a 1.3B backbone while generating each frame $9.5$--$28.5\times$ faster.
\end{itemize}
We review related work in Appendix~\ref{sec:related}.

\newcommand{\sdrift}{s}                 
\newcommand{\mubase}{\mu_{\mathrm{base}}}  
\newcommand{\bmag}{b}                   
\newcommand{\bcap}{b_{\max}}            
\newcommand{\gvec}{\mathbf{g}}          
\newcommand{\gtil}{\widetilde{g}}       
\newcommand{\Sbar}{\bar{S}}             
\newcommand{\Aset}{\mathcal{A}}         
\newcommand{\Rset}{\mathcal{R}}         
\newcommand{\Sset}{\mathcal{T}}         
\newcommand{\Lf}{L}                     
\newcommand{\Nslot}{N}                  
\newcommand{\enc}{\phi}                 
\newcommand{\method}{\textsc{Custom Forcing}}

\section{Preliminaries}
\label{sec:prelim}

\subsection{Streaming Autoregressive Video Diffusion}
\label{sec:prelim_ar}

Let a video be represented in the latent space of a causal VAE and divided into chunks $\mathbf{x} = (\mathbf{x}^1, \mathbf{x}^2, \dots)$, where each chunk $\mathbf{x}^t$ contains $F$ latent frames of $L = h \times w$ tokens. An autoregressive video diffusion model factorizes the distribution of the video over chunks,
\begin{equation}
    p_\theta(\mathbf{x} \mid y) \;=\; \prod_{t \ge 1} p_\theta\!\left(\mathbf{x}^t \mid \mathbf{x}^{<t}, y\right),
    \label{eq:ar_factorization}
\end{equation}
where $y$ is the text prompt, and samples each factor with a diffusion model. Because real-time generation leaves room for only a few network evaluations per chunk, recent models distill the denoiser into a few-step generator $G_\theta$~\citep{causvid,selfforcing}. Starting from $\mathbf{x}^t_K \sim \mathcal{N}(\mathbf{0}, \mathbf{I})$, chunk $t$ is obtained by
\begin{equation}
    \mathbf{x}^t_{k-1} \;=\; \Psi\!\left( G_\theta\big(\mathbf{x}^t_k, \sigma_k;\, y, \mathcal{C}_{t-1}\big),\; \sigma_{k-1} \right), \qquad k = K, \dots, 1,
    \label{eq:few_step}
\end{equation}
where $G_\theta$ predicts the clean chunk, $\Psi(\cdot, \sigma)$ re-noises its input to level $\sigma$ with $\sigma_0 = 0$, and $\mathbf{x}^t := \mathbf{x}^t_0$. The past enters only through $\mathcal{C}_{t-1}$, the KV cache that stores, for every transformer block, the keys and values of the frames finalized so far. Once $\mathbf{x}^t$ is complete, a forward pass on the clean chunk writes its keys and values into the cache, and $\mathbf{x}^t$ is never revised afterwards.

Inside each DiT block, a token of the chunk being denoised forms a query $\mathbf{q} \in \mathbb{R}^d$ and attends to the cached tokens together with the tokens of its own chunk. For a set $\mathcal{J}$ of token indices, we write
\begin{equation}
    \mathrm{Attn}(\mathbf{q}; \mathcal{J}) \;=\; \sum_{j \in \mathcal{J}} \frac{\exp\!\left(\mathbf{q}^\top \mathbf{k}_j / \sqrt{d}\right)}{\sum_{j' \in \mathcal{J}} \exp\!\left(\mathbf{q}^\top \mathbf{k}_{j'} / \sqrt{d}\right)} \, \mathbf{v}_j,
    \label{eq:attn}
\end{equation}
so that standard self-attention is $\mathrm{Attn}(\mathbf{q}; \mathcal{J}_{\mathcal{C}} \cup \mathcal{J}_{\mathrm{cur}})$, with $\mathcal{J}_{\mathcal{C}}$ indexing the cached tokens and $\mathcal{J}_{\mathrm{cur}}$ the tokens of the current chunk.

The cache cannot grow with the video, so streaming models bound it. A common design~\citep{rollingforcing,rollingsink,deepforcing} splits it as $\mathcal{C}_t = \mathcal{S} \cup \mathcal{W}_t$: an attention sink $\mathcal{S}$ that holds the first few frames and is kept for the entire rollout, and a window $\mathcal{W}_t$ of recent frames from which older frames are evicted. The sink is therefore the only part of the context that every chunk can see, and in the standard setting it holds the first frames that the model generates.

\subsection{Subject Customization in Autoregressive Diffusion}
\label{sec:prelim_setting}
We are given a small set of reference images $\{c_i\}_{i \in \mathcal{R}}$ of a particular subject and a prompt $y$ that describes a scene and an action but names the subject only by its category (e.g., ``a dog trotting down a forest trail''). The goal is to generate a video that follows $y$ and in which the subject matches the reference images throughout the rollout. The generator $G_\theta$ stays frozen, and nothing is optimized for the subject.

Let $s(t)$ denote the similarity between the subject in chunk $\mathbf{x}^t$ and the reference images, which we define concretely in Eq.~\eqref{eq:s}. Identity drift (Sec.~\ref{sec:intro}) is the decline of $s(t)$ over the rollout. It differs from temporal consistency, which compares $\mathbf{x}^t$ with the preceding chunks $\mathbf{x}^{<t}$, because a video can remain smooth and self-consistent while its subject slowly turns into a generic member of the category. Any correction must therefore be made online, before the next chunk is generated.

\section{\method{}: Subject Customization in the KV Cache}
\label{sec:method}

\subsection{Overview}
\label{sec:overview}

Existing video customization methods introduce a subject through per-subject optimization or an additional conditioning network (see Appendix~\ref{sec:related}). A autoregressive video model may need neither, because every chunk already reads the KV cache and part of the cache persists for the entire rollout. We hypothesize that images of the subject written into this persistent part can serve as its reference throughout generation if the model reads them with the right strength and in the right direction.

\begin{figure}[t]
\centering
    \includegraphics[width=1\linewidth]{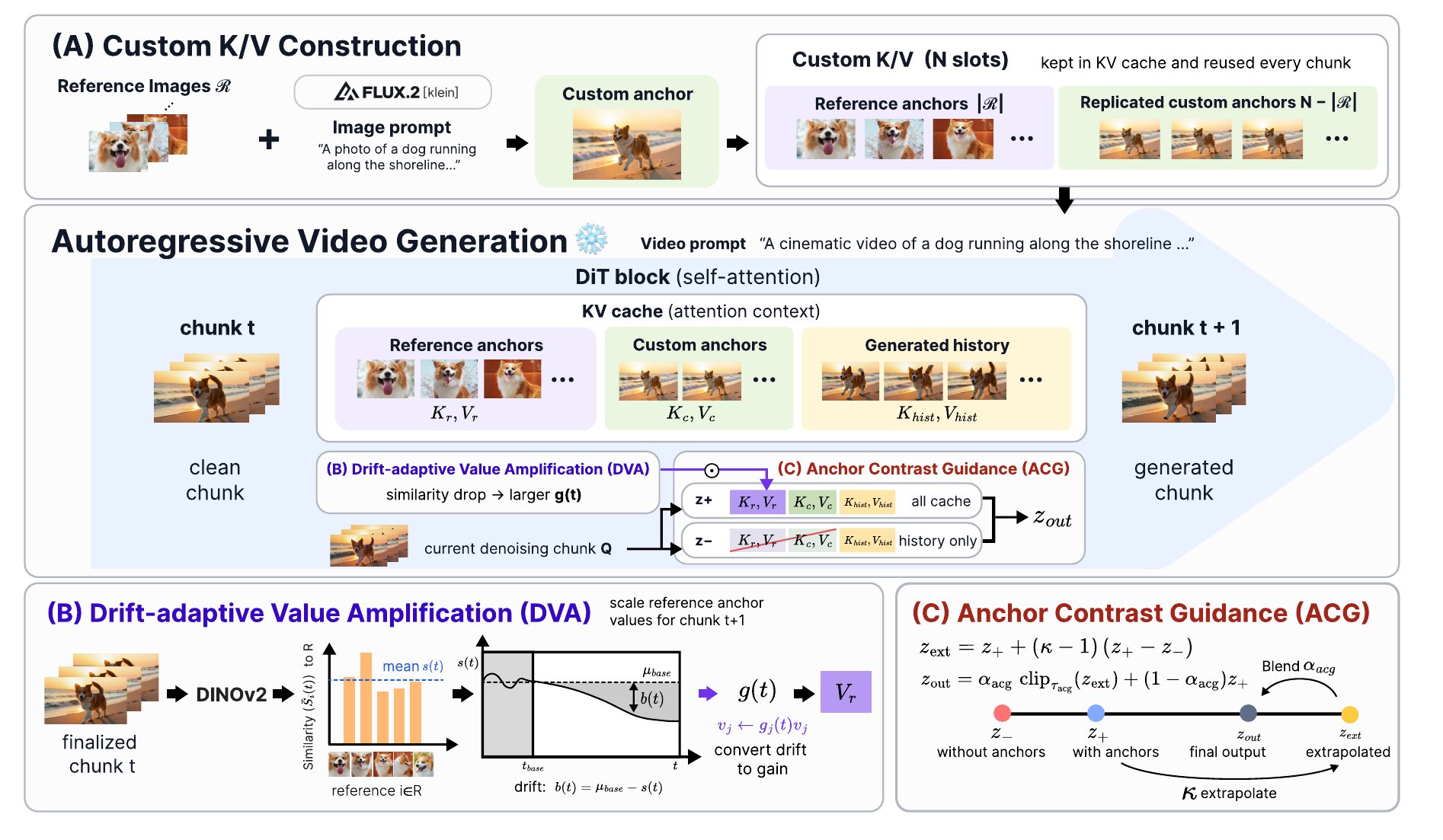}
    \caption{\textbf{Overview of \method{}.} \textbf{Top:} the reference images and copies of a custom anchor generated by FLUX.2\,[klein]~\citep{flux-2-2025} are written into the KV cache once as the custom K/V. \textbf{Middle:} in every self-attention layer of the frozen backbone, the current chunk reads the cache with and without the anchor frames, giving $z_{+}$ and $z_{-}$, after DVA scales the reference values by $g(t)$. \textbf{Bottom left:} DVA sets $g(t)$ from how far the similarity $\sdrift(t)$ of each finalized chunk falls below its initial level $\mubase$. \textbf{Bottom right:} ACG extrapolates away from $z_{-}$ by $\kappa$ and blends the result with $z_{+}$.}
    \label{fig:main_figure}
\end{figure}

\method{} has three components (Fig.~\ref{fig:main_figure}). Custom K/V construction (Sec.~\ref{sec:setup}) writes the reference images and one custom anchor of the subject into the cache before generation. Drift-adaptive value amplification (DVA, Sec.~\ref{sec:gate}) sets how strongly the reference images are read, and anchor contrast guidance (ACG, Sec.~\ref{sec:nag}) sets the direction away from the generic subject. DVA rescales the values before attention is computed, and ACG modifies the attention output.

\subsection{Custom K/V Construction}
\label{sec:setup}

Before the first chunk is generated, we encode $\Nslot$ images of the subject with the VAE, one latent frame per image, and pass them through $G_\theta$ as clean frames, which writes their keys and values into the cache of every block. We call these frames the \emph{custom K/V} and index them by $\Aset = \{1, \dots, \Nslot\}$. The custom K/V occupies the attention sink of Sec.~\ref{sec:prelim_ar} in place of the first generated frames. It precedes all generated frames and is never evicted, so every chunk reads it through the ordinary attention of Eq.~\eqref{eq:attn}, without a retrieval step or an additional conditioning branch.

\paragraph{Reference anchors specify the subject, and the custom anchor specifies the opening scene.}
The custom K/V contains two kinds of frames. The \emph{reference anchors}, at positions $\Rset \subset \Aset$, are the user's reference images used as provided; they determine what the subject looks like. The remaining positions $\Aset \setminus \Rset$ hold the \emph{custom anchor}, a single image of the subject placed in the scene described by $y$. We generate it once with FLUX.2\,[klein]~\citep{flux-2-2025} from the reference images and a short image prompt derived from $y$ (Appendix~\ref{app:impl}). Because the video begins from this image, we replicate it in all $\Nslot - |\Rset|$ remaining positions and place it last, directly before the first chunk.

More copies of the custom anchor keep the opening scene stable but fix the subject to its pose, whereas more reference anchors restore motion at the cost of scene stability (Appendix~\ref{app:rladder}).

\paragraph{The custom K/V alone customizes the video but cannot prevent drift.}
Writing the custom K/V into the cache and changing nothing else already produces a customized video, which we call \emph{Custom K/V only}. It cannot prevent identity drift, because it reads the anchors with a fixed strength and does nothing against the bias of the prompt toward a generic subject. Sec.~\ref{sec:gate} and~\ref{sec:nag} address these two limitations.

\subsection{Drift-Adaptive Value Amplification}
\label{sec:gate}

\paragraph{A fixed reference strength cannot serve the whole rollout.}
Reading the anchor frames more strongly reduces drift, but doing so throughout the video enlarges the subject and reduces motion (Appendix~\ref{app:const}). We therefore scale the reference anchors with the measured drift.

\paragraph{Drift can be measured online from finalized chunks.}
We decode each finalized chunk into $F'$ pixel frames $\{x_f(t)\}_{f=1}^{F'}$, where $F' > F$ because the VAE compresses time, and embed each frame with a frozen DINOv2-B encoder $\enc$~\citep{dinov2}, using the $\ell_2$-normalized \textsc{cls} token. The similarity of chunk $t$ to reference anchor $c_i$ and its average over the reference anchors are
\begin{equation}
\Sbar_i(t) \;=\; \frac{1}{F'} \sum_{f=1}^{F'} \big\langle \enc(x_f(t)),\, \enc(c_i) \big\rangle, \quad i \in \Rset,
\qquad
\sdrift(t) \;=\; \frac{1}{|\Rset|} \sum_{i \in \Rset} \Sbar_i(t).
\label{eq:s}
\end{equation}
We exclude the custom anchor, because the video starts from it.

The absolute value of $\sdrift(t)$ varies across subjects and prompts, so we compare it with a level estimated from the same video. The first $t_{\text{base}}$ chunks keep the values unchanged, and their mean similarity defines
\begin{equation}
\mubase \;=\; \frac{1}{t_{\text{base}}} \sum_{t=1}^{t_{\text{base}}} \sdrift(t).
\label{eq:mu0}
\end{equation}
From then on, drift is the amount by which $\sdrift(t)$ falls below this level, clipped to $[0, \bcap]$:
\begin{equation}
\bmag(t) \;=\; \min\!\Big( \max\!\big(0,\; \mubase - \sdrift(t)\big),\; \bcap \Big).
\label{eq:b}
\end{equation}

\paragraph{Amplification is restricted to the subject in each reference image.}
The drift is converted into an amplification of the values of the reference anchors, while the custom anchor, whose background is the opening scene, is left unchanged. Only the subject in a reference image carries identity, so we compute a subject mask for each reference anchor with SAM~\citep{sam} and average it within each latent token. Each token $j$ then carries a weight $w_j \in [0, 1]$, the fraction of its pixels that belong to the subject. Tokens with $w_j \le \tau_{\text{subj}}$ are treated as background and removed from attention, and the remaining tokens of reference anchor $i$ form the set $\Sset(i)$ with $n_i = |\Sset(i)|$. For $j \in \Sset(i)$ and $i \in \Rset$, the target amplification is
\begin{equation}
\gtil_j(t) \;=\; \frac{\alpha_{\text{inj}}\, \bmag(t)}{|\Rset|} \cdot \frac{n_i\, w_j}{\sum_{j' \in \Sset(i)} w_{j'}}.
\label{eq:gate}
\end{equation}
The first factor divides a total budget $\alpha_{\text{inj}}\, \bmag(t)$ evenly among the reference anchors, and the second has mean one over $\Sset(i)$, so it moves the share of each anchor toward the tokens that are mostly subject (Appendix~\ref{app:impl}).

Since $\bmag(t)$ can change abruptly between chunks, we smooth the amplification with an exponential moving average,
\begin{equation}
g_j(t) \;=\; \lambda\, g_j(t-1) + (1 - \lambda)\, \gtil_j(t),
\label{eq:ema}
\end{equation}
and read the values of these tokens as $g_j(t)\, \mathbf{v}_j$ for $j \in \bigcup_{i \in \Rset} \Sset(i)$, leaving the cached $\mathbf{v}_j$ unchanged. The amplification computed after chunk $t$ is finalized takes effect from chunk $t+1$ onward. We set $g_j(t) = 1$ for $t < t_{\text{base}}$, as in Custom K/V only, after which $g_j$ falls below $1$ for small drift and rises above $1$ as it grows.

\subsection{Anchor Contrast Guidance}
\label{sec:nag}

\paragraph{The prompt biases the subject toward a generic member of its category.}
DVA determines how strongly the anchor frames are read, but not what the subject is drifting toward. The prompt names only the category of the subject, so the backbone tends to produce a typical member of that category, which we call the \emph{class prior}. The subject of Custom K/V only drifts toward such generic instances over time (Fig.~\ref{fig:qual}). Counteracting the class prior requires a direction in feature space, and a second read of the same cache provides one.

\paragraph{Masking the anchors yields a negative branch from the same cache.}
Let $\mathcal{J}_{\Aset}$ index the anchor tokens and $\mathcal{J}_{\text{gen}}$ all other tokens a query can attend to, namely the retained generated frames and the chunk being denoised. Since the two sets are disjoint, the same query can be evaluated with and without the anchors:
\begin{equation}
z_{+} \;=\; \mathrm{Attn}(\mathbf{q};\, \mathcal{J}_{\Aset} \cup \mathcal{J}_{\text{gen}}),
\qquad
z_{-} \;=\; \mathrm{Attn}(\mathbf{q};\, \mathcal{J}_{\text{gen}}).
\label{eq:nagbranch}
\end{equation}
The output $z_{-}$ still reflects the anchor frames indirectly, because the generated frames in $\mathcal{J}_{\text{gen}}$ were produced under their influence. The two outputs differ only in whether this query attends to the anchor frames at the current step.

Since the softmax in Eq.~\eqref{eq:attn} factorizes over disjoint index sets~\citep{refdrop}, $z_{+}$ is a convex combination of the two parts:
\begin{equation}
z_{+} \;=\; m\, z_{\Aset} + (1 - m)\, z_{-},
\qquad
m \;=\; \frac{M_{\Aset}}{M_{\Aset} + M_{\text{gen}}} \in [0, 1],
\label{eq:acgpartition}
\end{equation}
where $z_{\Aset} = \mathrm{Attn}(\mathbf{q};\, \mathcal{J}_{\Aset})$ and $M_{\mathcal{J}} = \sum_{j \in \mathcal{J}} \exp(\mathbf{q}^\top \mathbf{k}_j / \sqrt{d})$. Hence $z_{+} - z_{-} = m\, (z_{\Aset} - z_{-})$. This difference points from the anchor-free output toward what the anchors contribute and is scaled by $m$, the share of attention that the query places on the anchors.

\paragraph{Extrapolating along this difference strengthens the anchors against the class prior.}
Following NAG~\citep{nag}, we extrapolate with strength $\kappa$,
\begin{equation}
z_{\text{ext}} \;=\; z_{+} + (\kappa - 1)\,(z_{+} - z_{-}),
\label{eq:acggate}
\end{equation}
which has no effect for $\kappa = 1$. To keep the extrapolated feature within a reasonable range, we bound its $\ell_1$ norm relative to $z_{+}$ and blend the result with $z_{+}$:
\begin{equation}
z_{\text{out}} \;=\; \alpha_{\text{acg}}\, \mathrm{clip}_{\tau_{\text{acg}}}(z_{\text{ext}}) + (1 - \alpha_{\text{acg}})\, z_{+},
\qquad
\mathrm{clip}_{\tau}(z) \;=\; \min\!\left(1,\, \tau\, \frac{\|z_{+}\|_1}{\|z\|_1}\right) z.
\label{eq:acgout}
\end{equation}
When the clip is inactive, substituting Eq.~\eqref{eq:acgpartition} into Eqs.~\eqref{eq:acggate} and~\eqref{eq:acgout} gives
\begin{equation}
z_{\text{out}} \;=\; m'\, z_{\Aset} + (1 - m')\, z_{-},
\qquad
m' \;=\; \big(1 + \alpha_{\text{acg}}(\kappa - 1)\big)\, m .
\label{eq:acgshare}
\end{equation}
ACG therefore multiplies the attention share that each query already places on the anchor frames, while the clip bounds the norm of $z_{\text{ext}}$. Unlike RefDrop~\citep{refdrop}, which replaces this query-dependent share with a constant, ACG scales it, so queries that already read the subject from the anchors read it more strongly. Unlike NAG~\citep{nag}, which contrasts a positive and a negative prompt in cross-attention, ACG obtains its negative branch by masking the anchor frames in self-attention. Neither branch requires an additional forward pass of the network (Appendix~\ref{sec:rel:control}).

ACG is applied to the self-attention of every DiT block at every denoising step, and cross-attention to the text is left unchanged. Like DVA, it starts after the first $t_{\text{base}}$ chunks, with a linear warm-up (Appendix~\ref{app:warmup}). DVA acts on the next chunk, whereas ACG acts on the current one.

\section{Experiments}
\label{sec:exp}

\providecommand{\cellcolor}[2][]{}
\definecolor{oursrow}{gray}{0.92}
\newcommand{\oc}{\cellcolor{oursrow}}

\subsection{Experimental Setup}
\label{sec:exp:setup}

\paragraph{Implementation details.}
We apply \method{} to Rolling Forcing~\citep{rollingforcing}, a streaming autoregressive video diffusion model built on Wan2.1-T2V-1.3B~\citep{wan}, without fine-tuning. We use $10$ DreamBooth subjects~\citep{dreambooth}, six animals and four objects, with $10$ prompts each Appendix~\ref{app:prompts} that name only the category of the subject, and generate one $30$\,s and one $2$\,min video per prompt. The custom K/V holds five reference anchors and four copies of one custom anchor, which FLUX.2\,[klein] 4B~\citep{flux-2-2025} generates for each prompt. We use $\alpha_{\text{inj}} = 300$ and $\kappa = 4$ for every subject and prompt (more details are in Appendix~\ref{app:impl}).

\paragraph{Baselines.}
We compare with the bidirectional customization methods SMRABooth~\citep{smrabooth}, CustomCrafter~\citep{customcrafter}, MotionBooth~\citep{motionbooth}, and the training-free VACE-1.3B~\citep{vace} at their native lengths of $2$--$5.1$\,s. We also compare with the image-to-video models FramePack~\citep{framepack}, FramePack-F1, and SkyReels-V2 DF~\citep{skyreelsv2}, which start from the custom anchor, and the reference-to-video model SkyReels-V3~\citep{skyreelsv3}. Finally, we report the \emph{vanilla AR model} with the prompt alone, \emph{vanilla AR + LoRA} fine-tuned on each subject, and \emph{Custom K/V only} without DVA or ACG (Appendix~\ref{app:impl}).

\paragraph{Evaluation metrics.}
DINO-I and CLIP-I measure similarity to the reference images, and CLIP-T measures agreement with the prompt. To expose drift, we also compute DINO-I on the first, middle, and last thirds of each $30$\,s video and on windows of the same duration at the start, middle, and end of each $2$\,min video. From VBench~\citep{vbench}, we report Subject and Background Consistency and Dynamic Degree. All scores are means over the $100$ prompts (see Appendix~\ref{app:eval}).

\begin{table}[t]
\centering
\small
\setlength{\tabcolsep}{2.0pt}
\renewcommand{\arraystretch}{1.12}
\resizebox{\linewidth}{!}{%
\begin{tabular}{@{}lccc cccc cc ccc@{}}
\toprule
& & & & \multicolumn{4}{c}{DINO-I over time $\uparrow$} & \multicolumn{2}{c}{Alignment $\uparrow$} & \multicolumn{3}{c}{Video quality} \\
\cmidrule(lr){5-8}\cmidrule(lr){9-10}\cmidrule(l){11-13}
Method & Length & \shortstack{Long\\video} & \shortstack{Training\\free} & all & first & middle & last & CLIP-I & CLIP-T & \shortstack{Subj.\\Cons.\,$\uparrow$} & \shortstack{Bg.\\Cons.\,$\uparrow$} & \shortstack{Dyn.\\Degree} \\
\midrule
\multicolumn{13}{@{}l}{\textit{Bidirectional}} \\
\quad VACE-1.3B & 5.1\,s & \xmark & \cmark & 0.333 & -- & -- & -- & 0.703 & 0.296 & 0.921 & 0.951 & 0.46 \\
\quad SMRABooth & 3.3\,s & \xmark & \xmark & 0.407 & -- & -- & -- & 0.664 & 0.309 & \textbf{0.978} & 0.969 & 0.21 \\
\quad CustomCrafter & 2.0\,s & \xmark & \xmark & 0.490 & -- & -- & -- & 0.669 & 0.301 & 0.966 & \textbf{0.971} & 0.17 \\
\quad MotionBooth & 3.0\,s & \xmark & \xmark & 0.435 & -- & -- & -- & 0.642 & 0.293 & 0.907 & 0.928 & 0.76 \\
\midrule
\multicolumn{13}{@{}l}{\textit{Autoregressive}} \\
\quad Vanilla AR model & 30\,s & \cmark & -- & 0.163 & 0.163 & 0.160 & 0.155 & 0.681 & \textbf{0.360} & 0.924 & 0.943 & 0.36 \\
\quad Vanilla AR + LoRA & 30\,s & \cmark & \xmark & 0.435 & 0.418 & 0.432 & 0.413 & 0.642 & 0.331 & 0.936 & 0.948 & 0.28 \\
\quad Custom K/V only & 30\,s & \cmark & \cmark & 0.541 & 0.579 & 0.506 & 0.475 & 0.791 & 0.355 & 0.913 & 0.939 & 0.41 \\
\oc\quad\method{} (Ours) & \oc30\,s & \oc\cmark & \oc\cmark & \oc\textbf{0.635} & \oc\textbf{0.619} & \oc\textbf{0.626} & \oc\textbf{0.616} & \oc\textbf{0.803} & \oc0.353 & \oc0.928 & \oc0.945 & \oc0.41 \\
\cmidrule(l){1-13}
\quad Custom K/V only & 2\,min & \cmark & \cmark & 0.501 & 0.583 & 0.450 & 0.415 & 0.785 & \textbf{0.357} & 0.882 & 0.925 & 0.47 \\
\oc\quad\method{} (Ours) & \oc2\,min & \oc\cmark & \oc\cmark & \oc\textbf{0.623} & \oc\textbf{0.622} & \oc\textbf{0.602} & \oc\textbf{0.578} & \oc\textbf{0.798} & \oc0.352 & \oc\textbf{0.909} & \oc\textbf{0.936} & \oc0.45 \\
\bottomrule
\end{tabular}}
\caption{\textbf{Main results.} \emph{Long video} marks whether a method can extend generation beyond a fixed length, and \emph{Training free} whether it customizes the subject without training any model on it. Bold marks the best value in each column among the bidirectional methods and the $30$\,s rows, and the better of the two $2$\,min rows. Dynamic Degree is not bolded. Subject and Background Consistency favor the short bidirectional videos, which change little.}
\label{tab:main}
\end{table}

\subsection{Main Results}
\label{sec:exp:main}
\label{sec:exp:long}

\begin{figure}[t]
\centering
\includegraphics[width=0.95\linewidth]{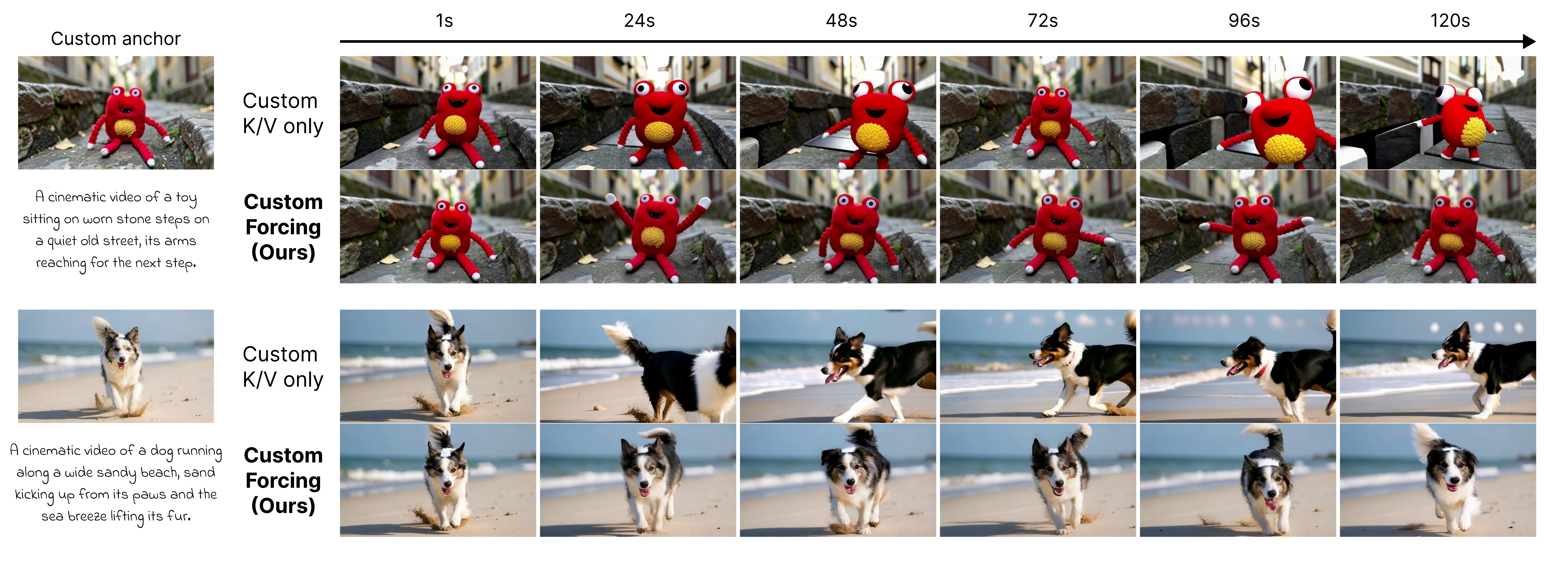}
\caption{\textbf{Two-minute generation.} Custom K/V only drifts within the first minute, while \method{} keeps both subjects for two minutes.}
\label{fig:long}
\end{figure}

\paragraph{Comparison with customization methods.}
Tab.~\ref{tab:main} gives the main results. The bidirectional methods, including the training-free VACE-1.3B, stay below Custom K/V only in DINO-I ($0.33$--$0.49$), and fine-tuning a LoRA of the backbone raises DINO-I from $0.163$ only to $0.435$, below the $0.541$ of Custom K/V only. Custom K/V only drifts from $0.579$ to $0.475$ within $30$\,s and from $0.583$ to $0.415$ over $2$\,min. \method{} stays nearly flat over $30$\,s ($0.619$ to $0.616$) and declines only to $0.578$ over $2$\,min, with Dynamic Degree at the level of Custom K/V only (Fig.~\ref{fig:long} and Appendix~\ref{app:main}).

\begin{table}[t]
\centering
\small
\setlength{\tabcolsep}{3pt}
\renewcommand{\arraystretch}{1.12}
\newcommand{\ox}{\cmark}
\newcommand{\xx}{\xmark}
\resizebox{\linewidth}{!}{%
\begin{tabular}{@{}ll c c cccc cc@{}}
\toprule
& & & & \multicolumn{4}{c}{DINO-I over time $\uparrow$} & \multicolumn{2}{c}{Alignment $\uparrow$} \\
\cmidrule(lr){5-8}\cmidrule(l){9-10}
Method & Backbone & Time & Causal & all & first & middle & last & CLIP-I & CLIP-T \\
\midrule
FramePack & HunyuanVideo 13B & 9.4\texttimes & \xx & 0.625 & 0.616 & \underline{0.620} & \textbf{0.623} & \underline{0.795} & \underline{0.357} \\
FramePack-F1 & HunyuanVideo 13B & 9.5\texttimes & \ox & 0.477 & 0.534 & 0.418 & 0.392 & 0.773 & 0.355 \\
SkyReels-V2 DF & Wan2.1 1.3B & 11.5\texttimes & \ox & 0.449 & 0.552 & 0.393 & 0.307 & 0.767 & \textbf{0.361} \\
SkyReels-V3 & Wan2.1 14B & 28.5\texttimes & \ox & \textbf{0.641} & \textbf{0.693} & \underline{0.620} & 0.564 & 0.794 & 0.347 \\
\oc\method{} (Ours) & \oc Wan2.1 1.3B & \oc1.0\texttimes & \oc\ox & \oc\underline{0.635} & \oc\underline{0.619} & \oc\textbf{0.626} & \oc\underline{0.616} & \oc\textbf{0.803} & \oc0.353 \\
\bottomrule
\end{tabular}}
\caption{\textbf{Comparison with image- and reference-to-video models}, $30$\,s, $n=100$, identity and prompt alignment only. \emph{Backbone} gives the architecture and size of each video diffusion transformer, \emph{Time} is the generation time per frame relative to \method{} on one H100 GPU, and \emph{Causal} marks whether each part of the video is final before the next is generated. Bold marks the best score and underline the second best.}
\label{tab:i2v}
\end{table}

\begin{figure}[t]
\centering
\includegraphics[width=1\linewidth]{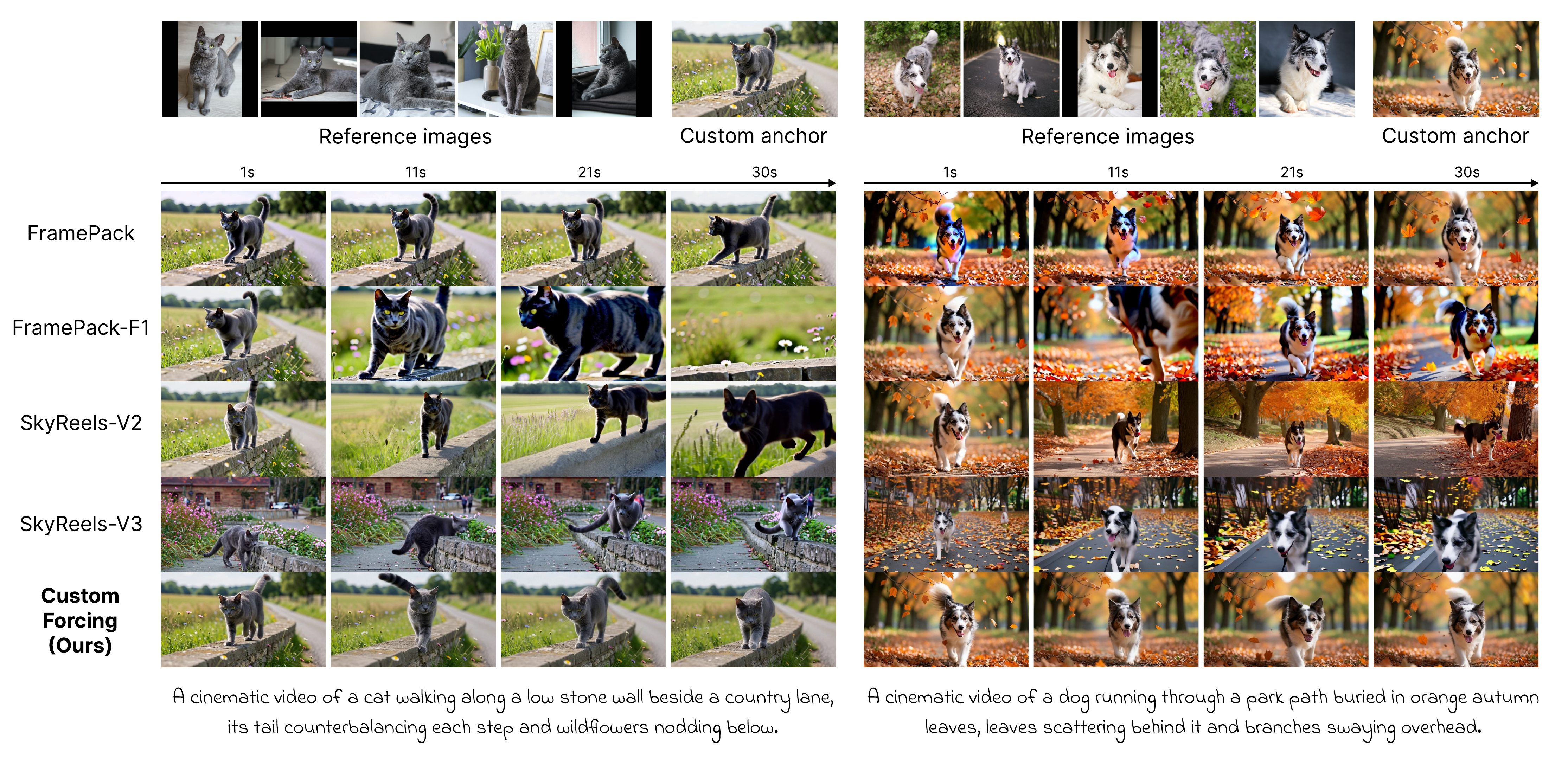}
\caption{\textbf{Qualitative comparison with image- and reference-to-video models.} The image-to-video models start from the custom anchor, and SkyReels-V3 starts from the reference images, so its scenes differ. \method{} keeps both subjects to the end.}
\label{fig:i2v}
\end{figure}

\paragraph{Comparison with image- and reference-to-video models.}
Tab.~\ref{tab:i2v} compares \method{} with models that start from the custom anchor or the reference images, a two-stage route that prior customization work also uses as a baseline~\citep{sugar,consisid}. With a 1.3B backbone, \method{} matches the 14B SkyReels-V3 in whole-video DINO-I ($0.635$ against $0.641$) and exceeds it in the last window ($0.616$ against $0.564$), while generating each frame $9.4$--$28.5\times$ faster than all four models. FramePack does not decline but scores lower overall and must fix the video length before generation, and the causal FramePack-F1 and SkyReels-V2 DF drift within $30$\,s (Fig.~\ref{fig:i2v}). A user study with $30$ participants prefers \method{} over every other method it includes except SkyReels-V3, against which the difference is not significant (see Appendix~\ref{app:user}).

\paragraph{Inference cost.}
On one H100 GPU, \method{} takes $1.13\times$ the generation time of Custom K/V only at $30$\,s and $1.11\times$ at $2$\,min. Most of the overhead comes from measuring drift, and ACG alone adds $4\%$.

\subsection{Ablation Study}
\label{sec:exp:comp}

\begin{wraptable}{r}{0.40\linewidth}
\vspace{-12pt}
\centering
\scriptsize
\setlength{\tabcolsep}{2pt}
\renewcommand{\arraystretch}{1.05}
\begin{tabular}{@{}lcc ccc@{}}
\toprule
& & & \multicolumn{3}{c}{DINO-I $\uparrow$} \\
\cmidrule(l){4-6}
& DVA & ACG & first & mid. & last \\
\midrule
Custom K/V only & \xmark & \xmark & 0.579 & 0.506 & 0.475 \\
+\,DVA & \cmark & \xmark & 0.600 & 0.598 & 0.598 \\
\oc +\,DVA\,+\,ACG (Ours) & \oc\cmark & \oc\cmark & \oc\textbf{0.619} & \oc\textbf{0.626} & \oc\textbf{0.616} \\
\bottomrule
\end{tabular}
\caption{\textbf{Component ablation}, $30$\,s.}
\label{tab:comp}
\vspace{-12pt}
\end{wraptable}

\begin{figure}[t]
\centering
\includegraphics[width=1\linewidth]{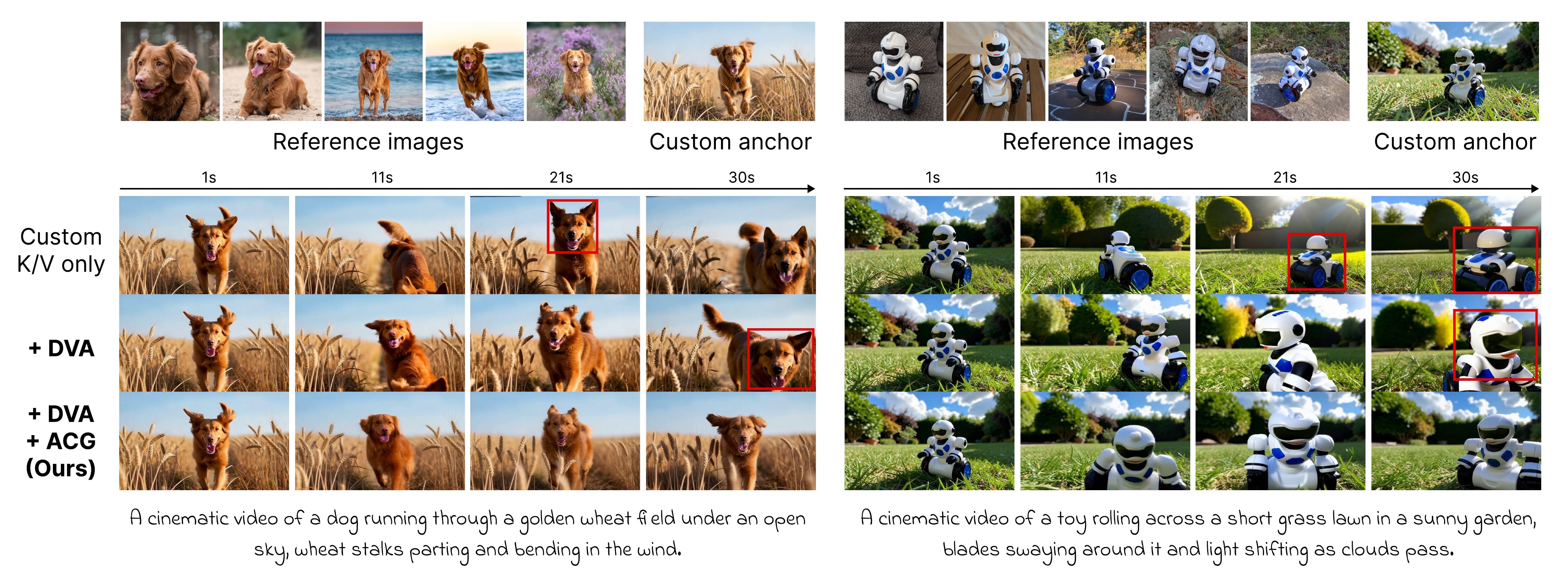}
\caption{\textbf{Adding the components one at a time.} Red boxes mark where the subject departs from the reference images.}
\label{fig:qual}
\end{figure}

\paragraph{Effect of DVA and ACG.}
Tab.~\ref{tab:comp} and Fig.~\ref{fig:qual} add the components one at a time. Adding DVA stops the drift, keeping DINO-I near $0.60$ from the first window to the last, and adding ACG raises every window by a further $0.02$--$0.03$. In Fig.~\ref{fig:qual}, Custom K/V only turns both subjects toward generic instances of their categories by $21$\,s. DVA delays this change but still alters the face and shoulders of the toy, and adding ACG keeps both subjects as in the reference images (see Appendix~\ref{app:qual}).

\paragraph{Further analyses.}
Replacing the DINOv2-B drift encoder of DVA with ConvNeXt-B~\citep{convnext} changes the scores by at most $0.003$, so the improvement does not come from sharing the encoder with the metric (Appendix~\ref{app:sensor}). ACG lowers the similarity to the vanilla video of the same prompt, a measure of the class prior, by $0.015$--$0.040$ in every third of the video (Appendix~\ref{app:prior}). DVA stops the decline of DINO-I after about $10$\,s, while ACG alone raises the starting identity but still declines, so the two address different failures (Appendix~\ref{app:time}). Participants show no significant preference in quality between \method{} and Custom K/V only ($45.8\%$, Appendix~\ref{app:user}). Appendix~\ref{app:ablations} ablates the remaining design choices.

\section{Conclusion}
\label{sec:conclusion}

In this paper, we propose \method{}, a training-free method that customizes autoregressive video generation with images of a user-provided subject through the KV cache. It needs neither per-subject optimization nor an additional conditioning network, only five reference images and one generated image of the subject in the scene, and it adds $11$--$13\%$ to the generation time of fixed anchors. The custom K/V, written into the cache once, establishes the identity from the first chunk. DVA reads the reference images more strongly the further the subject drifts, and ACG extrapolates away from the attention output computed without the anchor frames, steering generation away from the generic subject favored by the prompt. Without modifying the backbone, \method{} keeps the subject close to the reference images for two minutes without reducing motion, holding DINO-I between $0.58$ and $0.62$ while fixed anchors fall to $0.42$. Because it changes only what the KV cache holds and how it is read, it also applies to Deep Forcing, another autoregressive model with an attention sink (Appendix~\ref{app:deep}). Appendix~\ref{app:limit} discusses its limitations. A direction for future work is to keep a user's subject in interactive world models, which also generate video causally over long horizons.

\FloatBarrier

\subsection*{Ethics statement}
The user study in Appendix~\ref{app:user} asked $30$ participants to compare pairs of generated videos of animals and objects. \method{} places a specific subject from a few images into generated videos without training. Applied to images of real people without their consent, this ability could produce misleading or harmful videos. Our experiments use only the animals and objects of the public DreamBooth dataset~\citep{dreambooth}, and deployments of \method{} should obtain consent for the depicted subject and label generated videos as synthetic.

\bibliography{custom_forcing_vB}
\bibliographystyle{iclr2027_conference}

\clearpage

\appendix

The Appendix is organized as follows:
\begin{itemize}
\item Sec.~\ref{sec:related}: related work.
\item Sec.~\ref{app:impl}: the backbone and inputs, hyperparameters, the custom anchor, and the settings of the baselines.
\item Sec.~\ref{app:eval}: how the metrics are computed.
\item Sec.~\ref{app:more}: details of the main results, results on Deep Forcing, results per subject, the drift sensor, identity over time, and a user study with VLM judges.
\item Sec.~\ref{app:qual}: qualitative observations on the two-minute videos, the image- and reference-to-video models, and the component ablation.
\item Sec.~\ref{app:ablations}: injection and guidance strength, the composition of the custom K/V, generated images in place of the reference images, and the adaptive amplification and subject mask.
\item Sec.~\ref{app:limit}: limitations.
\end{itemize}

\section{Related Work}
\label{sec:related}

\subsection{Autoregressive video generation}
\label{sec:rel:ar}

Bidirectional video generation models~\citep{videoworldsimulators2024,wan} generate fixed-length videos by denoising the sequence jointly. Autoregressive alternatives produce frames or chunks causally, making it possible to stream outputs and extend generation beyond a fixed video length. Diffusion Forcing introduced per-frame noise levels for causal sequence generation~\citep{diffusionforcing}; CausVid and Self Forcing distilled or post-trained causal few-step generators from pretrained video diffusion models~\citep{causvid,selfforcing}; and MAGI-1 scaled chunk-wise autoregressive generation to long contexts~\citep{magi1}. Later work improves distillation and rollout training~\citep{apt2,causalforcing,causalforcingpp,rewardforcing,streamforcing,resamplingforcing}, reduces attention and KV-cache cost~\citep{framepack,lightforcing,sparseforcing,forcingkv}, or trains causal world models for interactive control~\citep{nextforcing,arforcing,actionparty}.

Long rollouts accumulate errors beyond the training horizon. Recent methods preserve selected parts of the generated history through rolling attention sinks~\citep{rollingforcing,rollingsink,deepforcing,dysink,relaxforcing}, positional re-indexing, anchor tokens, and evolving memory states~\citep{anchorforcing,memrope,infinityrope,adastate}, longer context windows~\citep{selfforcingpp,longlive,contextforcing,packforcing}, or explicit spatial memory~\citep{memoryforcing}. Bagger and Pathwise Test-Time Correction address rollout error through backward aggregation or test-time correction from an initial frame~\citep{bagger,pathwise}, while IAMFlow tracks identities that have already appeared in an evolving generated narrative~\citep{iamflow}. These methods improve consistency with generated context. Custom Forcing instead maintains fidelity to user-provided reference images that are external to the rollout.

\subsection{Subject-driven video customization}
\label{sec:rel:video}

Subject customization has been extensively studied for images. Textual Inversion, DreamBooth, and Custom Diffusion optimize embeddings or model parameters for each subject~\citep{textualinversion,dreambooth,customdiffusion}, while ELITE, IP-Adapter, PhotoMaker, and InstantID learn reusable encoders or adapters for zero-shot conditioning~\citep{elite,ipadapter,photomaker,instantid}. These approaches establish the two main strategies used by later video methods: subject-specific optimization and pretrained reference conditioning.

Video customization methods based on subject-specific optimization include Magic-Me, DreamVideo, and CustomCrafter~\citep{magicme,dreamvideo,customcrafter}. CustomVideo instead combines attention control with subject masks for multi-subject customization~\citep{customvideo}. Still-Moving transfers an image-customized model to video without requiring customized video data~\citep{stillmoving}. A second group uses pretrained or reusable video-conditioning systems for zero-shot or feed-forward customization, including VideoBooth, ID-Animator, ConsisID, VideoMaker, SUGAR, MagicMirror, and Stand-In~\citep{videobooth,idanimator,consisid,videomaker,sugar,magicmirror,standin}. Recent systems extend this setting to multiple subjects, broader control signals, or unified creation interfaces~\citep{conceptmaster,movieweaver,phantom,skyreelsa2,hunyuancustom,vace,videoalchemist}. These methods primarily target bidirectional video generation and typically denoise a bounded video jointly. They generally do not address open-ended causal rollouts, where completed frames cannot be revised and reference information must remain available throughout generation.

V-Warper is closely related because it injects reference appearance through value representations without additional video training~\citep{vwarper}. Its coarse stage still uses image-domain subject adaptation, and its fine stage operates during bidirectional video denoising. Custom Forcing instead stores reference images in the persistent cache of a streaming autoregressive model and adapts their contribution from the observed rollout.

\subsection{Inference-time attention control}
\label{sec:rel:control}

Classifier-free guidance combines conditional and unconditional model predictions at inference time~\citep{cfg}. Normalized Attention Guidance (NAG) instead extrapolates between the attention outputs for a positive and a negative prompt and normalizes the result, which keeps negative guidance stable in few-step sampling~\citep{nag}. Prompt-to-Prompt and MasaCtrl manipulate cross-attention or self-attention to control image editing and consistent synthesis~\citep{prompttoprompt,masactrl}. For consistent subjects without training, ConsiStory and StoryDiffusion share self-attention across the images of a batch~\citep{consistory,storydiffusion}, and RefDrop shows that such sharing interpolates self-attention and reference attention, replacing the interpolation weight with a user-set constant~\citep{refdrop}. Custom Forcing builds on these ideas for a causal stream. ACG adopts the normalization of NAG but derives its negative branch by masking reference frames in the cache rather than from a negative prompt, and, unlike RefDrop, it scales the attention share that each query already places on the references instead of mixing in a constant amount, so queries that place little attention on the references remain nearly unchanged. Unlike classifier-free guidance, neither branch of ACG requires an additional forward pass, only a second attention read in each block. DVA further adjusts the reference values during sampling from an online measurement of subject similarity.

\section{Implementation Details}
\label{app:impl}
\label{app:hp}
\label{app:warmup}

\paragraph{Backbone and inputs.}
Rolling Forcing generates $832\times480$ video at $16$ fps in chunks of $F = 3$ latent frames, and each query attends to $27$ frames, namely the $\Nslot = 9$ anchor frames, $H = 3$ recent frames, and the $15$ frames being denoised. The custom K/V holds the first five DreamBooth images of each subject as $|\Rset| = 5$ reference anchors, followed by $4$ copies of the custom anchor. Rolling Forcing keeps several chunks in flight at staggered noise levels, which changes when a chunk is finalized but not the role of the cache. In Eq.~\eqref{eq:gate}, the second factor keeps the average amplification per token of a reference anchor independent of how much of the image the subject covers. We set $\alpha_{\text{inj}} = 300$ and $\kappa = 4$, which give the best trade-off between identity and motion (Appendix~\ref{app:alphak}).

\paragraph{Hyperparameters.}
Tab.~\ref{tab:hp} lists every constant in \method{}. After $\mubase$ is set, the ACG blend weight increases linearly to its final value $\alpha_{\text{acg}}$ over $W_{\text{warm}}$ chunks:
\begin{equation*}
\alpha_{\text{acg}}(t) \;=\; \alpha_{\text{acg}} \cdot \min\!\Big(1,\; \max\Big(0,\; \frac{t - t_{\text{base}}}{W_{\text{warm}}}\Big)\Big).
\end{equation*}
We generate every video with seed $0$. The nine images of the custom K/V take consecutive temporal positions before the first chunk, with the copies of the custom anchor last. The subject masks of the reference images come from SAM~\citep{sam}.

\begin{table}[h]
\centering
\small
\begin{tabular}{llc}
\toprule
 & Symbol & Value \\
\midrule
Anchor slots & $\Nslot$ & $9$ \\
Reference anchors & $|\Rset|$ & $5$ \\
Retained history frames & $H$ & $3$ \\
Chunk size / tokens per frame & $F$ / $\Lf$ & $3$ / $1560$ \\
\midrule
Injection budget & $\alpha_{\text{inj}}$ & $300$ \\
Drift cap & $\bcap$ & $0.2$ \\
Subject-mask threshold & $\tau_{\text{subj}}$ & $0.5$ \\
Baseline chunks & $t_{\text{base}}$ & $3$ \\
Amplification EMA & $\lambda$ & $0.95$ \\
\midrule
ACG extrapolation & $\kappa$ & $4$ \\
ACG blend & $\alpha_{\text{acg}}$ & $0.25$ \\
ACG clip & $\tau_{\text{acg}}$ & $2.5$ \\
ACG warm-up & $W_{\text{warm}}$ & $5$ chunks \\
\bottomrule
\end{tabular}
\caption{\textbf{Hyperparameters of \method{}.} No value is tuned per subject or per prompt, and only $\bmag(t)$ adapts to each video.}
\label{tab:hp}
\end{table}

\begin{figure}[t]
\centering
\includegraphics[width=\linewidth]{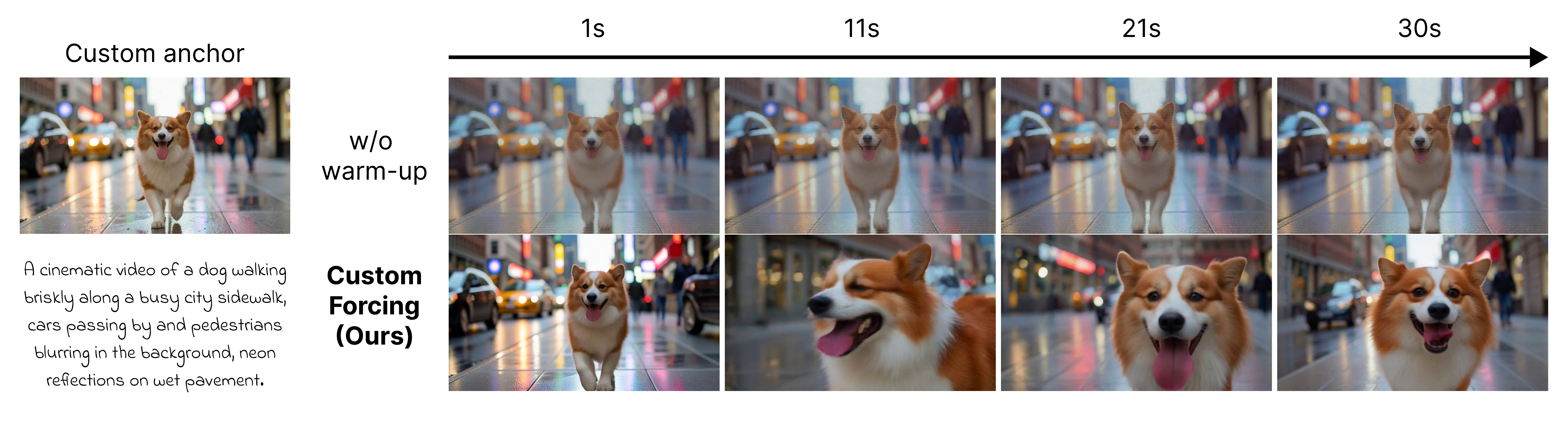}
\caption{\textbf{Effect of the ACG warm-up on Deep Forcing}, Without warm-up,
the generation remains close to the custom-anchor pose and exhibits
limited motion. With warm-up, the subject shows greater motion while preserving its identity.}
\label{fig:deepforcing_warmup}
\end{figure}

\paragraph{ACG warm-up.}
ACG is inactive during the first $t_{\mathrm{base}}$ chunks.
After $\mu_{\mathrm{base}}$ is established, we linearly increase
the blend weight $\alpha_{\mathrm{acg}}$ to its final value over
$W_{\mathrm{warm}}=5$ chunks, while keeping $\kappa$ fixed.
Fig.~\ref{fig:deepforcing_warmup} shows the qualitative effect
of this warm-up on Deep Forcing.

\paragraph{Custom anchor.}
FLUX.2\,[klein] 4B~\citep{flux-2-2025} receives the five reference images and an image prompt, which is the video prompt with its opening ``A cinematic video of'' replaced by ``A photo of''. It generates one $832\times480$ image with seed $0$, which we use without editing.

\paragraph{Bidirectional customization baselines.}
VACE-1.3B uses the released Wan2.1-VACE-1.3B checkpoint, which adds $15$ VACE context blocks to the same Wan2.1 1.3B transformer as our backbone, and the official Wan2.1 inference script. CustomCrafter is trained for $200$ epochs ($10$k optimizer steps at an effective batch size of $4$ and a learning rate of $3\times10^{-5}$) on the five reference images and $200$ class regularization images, following the official recipe. It samples $16$ frames at $320\times512$ and $8$ fps with $50$ DDIM steps, a guidance scale of $12$, and dynamic weighted sampling ($K=5$, $\lambda_s=0.4$, $\lambda_l=0.8$). Training for $50$ epochs instead lowers DINO-I from $0.490$ to $0.458$ but raises Dynamic Degree from $0.17$ to $0.32$, so longer training gains identity at the cost of motion. We report the $200$-epoch model, which gives CustomCrafter its highest identity.

SMRABooth customizes Wan2.1-T2V-1.3B with a spatial subject LoRA of rank $32$ on the query, key, and first feed-forward projections, regularized by representation alignment to DINOv2-ViT-g features with weight $0.05$. It is trained on the five reference images, their SAM masks, and the caption ``A picture of sks [class]'' at $512\times512$ for $2{,}000$ steps with batch size $1$, using AdamW with a constant learning rate of $10^{-4}$ in bf16. The official recipe uses two GPUs for $1{,}000$ steps at batch size $2$, so both see the same number of samples. SMRABooth samples $49$ frames at $480\times832$ and $15$ fps with $50$ steps and the official negative prompt, and raises the LoRA scale from $0.6$ to $0.7$ at step $25$.

MotionBooth fine-tunes the full UNet of Zeroscope v2 576w with its official configuration: a masked subject loss, a subject-token cross-attention loss with weight $0.01$, and a video loss with weight $1$. It is trained for $300$ steps at an effective batch size of $20$ with AdamW (learning rate $5\times10^{-6}$, weight decay $0.01$) and gradient clipping at $1$ in bf16. The Panda-70M clips that the official recipe uses for the video loss are no longer available, so we use $500$ clips of $24$ frames from UCF-101 instead. MotionBooth samples $24$ frames at $320\times576$ and $8$ fps with a guidance scale of $7.5$, without subject motion or camera control. For three subjects, it uses its own class names, ``toy car'', ``toy robot'', and ``toy monster''.

\paragraph{Training-based autoregressive baseline.}
Vanilla AR + LoRA fine-tunes Rolling Forcing itself. Following DreamBooth~\citep{dreambooth}, we add a new token \texttt{<sks>} to the text encoder and train it together with LoRA adapters of rank $16$ ($\alpha=16$) on the query, key, value, and output projections of the self- and cross-attention layers. Each reference image is repeated into a static clip of $9$ frames ($3$ latent frames) and flipped horizontally with probability $0.5$. The loss is the mean squared error of the predicted clean latent at one of the five denoising steps of the distilled generator, sampled uniformly and computed through the KV cache, plus a prior-preservation loss with weight $1$ on three clips from each class video of the vanilla AR model. We train for $1{,}000$ steps with AdamW, using a learning rate of $3\times10^{-4}$ and weight decay $0.01$ for the LoRA weights and a learning rate of $10^{-2}$ without weight decay for the token embedding, with $20$ warm-up steps, cosine decay, and gradient clipping at $1$. Training takes about $12$ minutes per subject. At inference, the LoRA scale is $0.5$.

\paragraph{Image- and reference-to-video baselines.}
All of them also receive the video prompt. FramePack and FramePack-F1 use their released checkpoints, which are built on HunyuanVideo, with $25$ sampling steps, distilled guidance $10$, and TeaCache disabled, and generate $832\times480$ video at $30$ fps. SkyReels-V2 DF uses the 1.3B 540P checkpoint in synchronous mode with $30$ sampling steps and guidance scale $6$, and generates $960\times544$ video at $24$ fps. FramePack-F1 is the forward-only variant of FramePack. SkyReels-V3 uses the R2V-14B checkpoint with the first four reference images, its maximum, and its official sampling settings, and generates $5$\,s of $1280\times720$ video at $24$ fps, which its official single-shot extension (V2V-14B) continues to $32.7$\,s. Its pipeline sets the output shape from the aspect ratio of the first reference image, so we pad that image to $16{:}9$ with the pipeline's own resizing. \method{} receives the same custom anchor and all five reference images. We cut every output to the length of our videos. The \emph{Time} column of Tab.~\ref{tab:i2v} averages the generation time per frame over the $100$ videos on one H100 GPU, covering every generation stage and VAE decoding but not model loading. Most of the overhead of \method{} over Custom K/V only comes from measuring drift, which decodes and embeds each finalized chunk.

\section{Evaluation Details}
\label{app:eval}

\paragraph{Identity and prompt alignment.}
We use every eighth frame of a video. DINO-I averages the DINOv2-B/14~\citep{dinov2} features of these frames, normalizes the mean, and takes its cosine similarity with each reference image, averaged over the five reference images. CLIP-I repeats this with CLIP ViT-B/32~\citep{clip} image features, and CLIP-T compares the same normalized mean with the CLIP text feature of the prompt. DINO-I over time applies the same computation to three $10.5$\,s windows. Our short videos have $501$ frames at $16$ fps ($31.3$\,s), which we round to $30$\,s in the text, so these windows are their three thirds.

\paragraph{VBench.}
Subject Consistency, Background Consistency, and Dynamic Degree come from the official VBench implementation~\citep{vbench}, run on each video at its native frame rate. Subject and Background Consistency measure temporal stability and Dynamic Degree measures motion, so together they show whether stability comes at the cost of motion.

\section{Additional Quantitative Results}
\label{app:more}

\subsection{Main Results in Detail}
\label{app:main}
The bidirectional methods also stay below Custom K/V only in CLIP-I and CLIP-T in Tab.~\ref{tab:main}, including VACE-1.3B, which adds context blocks to our Wan2.1 1.3B backbone. Their short videos change little, which gives SMRABooth and CustomCrafter the highest Subject and Background Consistency. Vanilla AR + LoRA lowers CLIP-I, CLIP-T, and Dynamic Degree below those of the vanilla AR model. \method{} raises both consistency scores over Custom K/V only, and its CLIP-T stays within $0.007$ of the highest value at $30$\,s, which the vanilla AR model reaches by following the prompt with no reference images. In Fig.~\ref{fig:long}, both subjects drift within the first minute with Custom K/V only, while \method{} keeps them for two minutes (Appendix~\ref{app:qual}).

In Tab.~\ref{tab:i2v}, the CLIP-T of \method{} stays within $0.01$ of the four image- and reference-to-video models. FramePack generates in inverted temporal order starting from the last window, so its DINO-I does not decline toward the end ($0.616$ to $0.623$), but its whole-video DINO-I is lower than ours ($0.625$ against $0.635$). Within $30$\,s, the DINO-I of FramePack-F1 falls from $0.534$ to $0.392$ and that of SkyReels-V2 DF from $0.552$ to $0.307$.

\subsection{Deep Forcing}
\label{app:deep}
We repeat the main experiments on Deep Forcing~\citep{deepforcing}, a training-free method that extends Self Forcing~\citep{selfforcing} to long videos. We keep its default participative compression (budget $16$, recent $4$), write the same custom K/V into its cache, and use the constants of Tab.~\ref{tab:hp} without change. Tab.~\ref{tab:deep} shows the same pattern as Tab.~\ref{tab:main}. Custom K/V only again does not prevent identity drift, and \method{} raises DINO-I from $0.556$ to $0.641$ at $30$\,s and from $0.518$ to $0.620$ at $2$\,min, with the largest improvement in the last window ($+0.110$ and $+0.125$). Adding DVA raises DINO-I in every window, and adding ACG on top gives the highest value in every window. On this backbone, however, DVA alone lowers Dynamic Degree from $0.81$ to $0.62$, and adding ACG restores it. At both lengths, Subject and Background Consistency stay at the level of Custom K/V only, and Dynamic Degree is at least as high ($0.87$ against $0.81$ at $30$\,s and $0.86$ against $0.84$ at $2$\,min). \method{} takes $1.30\times$ the time of Custom K/V only at $30$\,s and $1.26\times$ at $2$\,min.

\begin{figure}[t]
\centering
\makebox[\linewidth][l]{\small\textbf{(a)}~$30$\,s}\\[2pt]
\includegraphics[width=\linewidth]{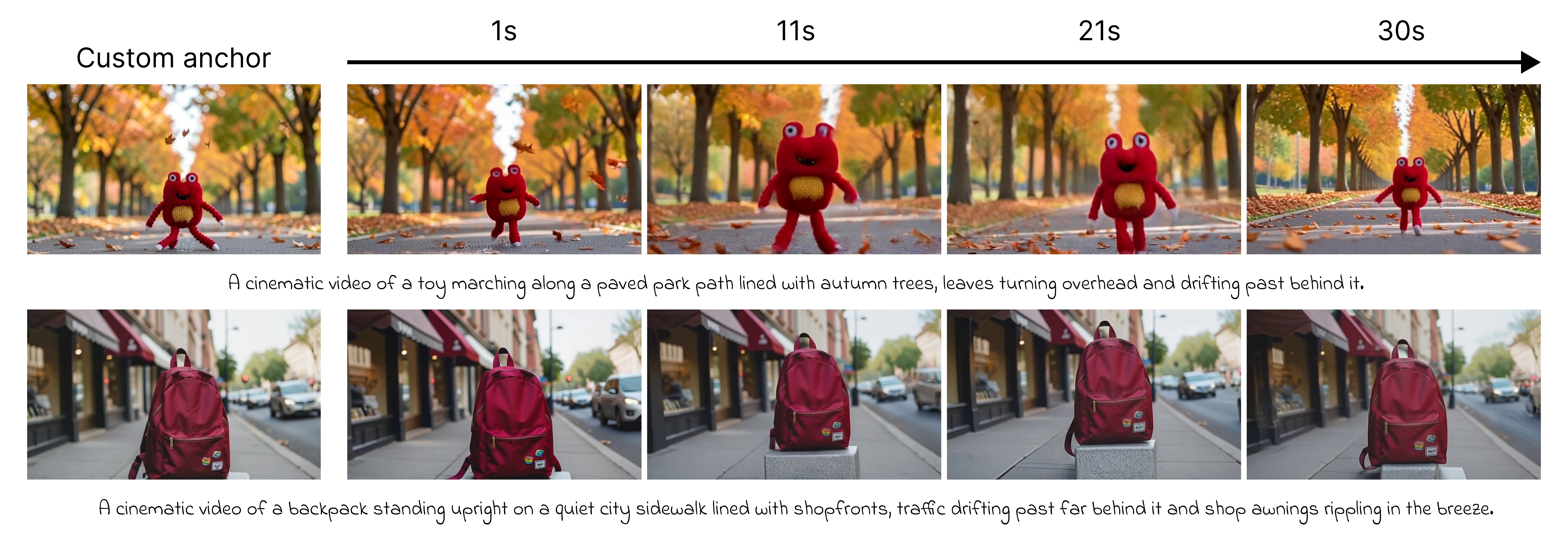}\\[8pt]
\makebox[\linewidth][l]{\small\textbf{(b)}~Two minutes}\\[2pt]
\includegraphics[width=\linewidth]{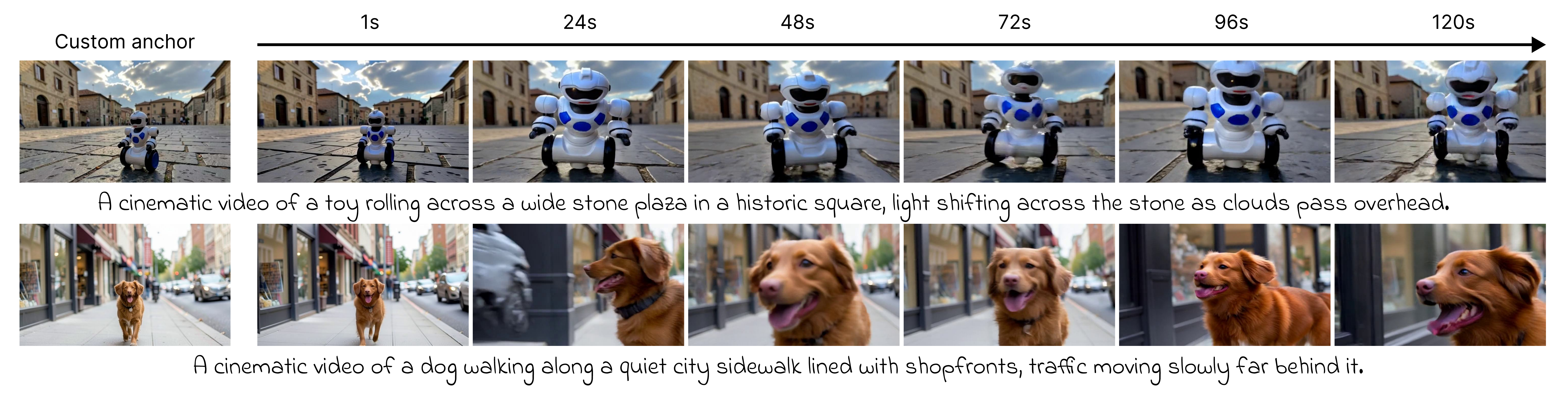}
\caption{\textbf{\method{} on Deep Forcing.} The custom anchor is on the left of each row. (a)~$30$\,s. Top: ``A toy marching along a paved park path lined with autumn trees, \ldots''. Bottom: ``A backpack standing upright on a quiet city sidewalk lined with shopfronts, \ldots''. (b)~Two minutes. Top: ``A toy rolling across a wide stone plaza in a historic square, \ldots''. Bottom: ``A dog walking along a quiet city sidewalk lined with shopfronts, \ldots''. \method{} keeps the red body, yellow belly, and large eyes of the toy and the red color and shape of the backpack to the end of the $30$\,s videos, and the white body, black visor, and blue circles of the robot and the red-brown coat and drooping ears of the dog for two minutes.}
\label{fig:deep_qual}
\end{figure}

Figure~\ref{fig:deep_qual} shows \method{} on Deep Forcing in $30$\,s and two-minute videos.

\begin{table}[h]
\centering
\small
\setlength{\tabcolsep}{3pt}
\renewcommand{\arraystretch}{1.1}
\resizebox{\linewidth}{!}{%
\begin{tabular}{@{}lccc cccc cc ccc@{}}
\toprule
& & & & \multicolumn{4}{c}{DINO-I over time $\uparrow$} & \multicolumn{2}{c}{Alignment $\uparrow$} & \multicolumn{3}{c}{Video quality} \\
\cmidrule(lr){5-8}\cmidrule(lr){9-10}\cmidrule(l){11-13}
Method & Length & DVA & ACG & all & first & middle & last & CLIP-I & CLIP-T & \shortstack{Subj.\\Cons.\,$\uparrow$} & \shortstack{Bg.\\Cons.\,$\uparrow$} & \shortstack{Dyn.\\Degree} \\
\midrule
Vanilla AR model & 30\,s & -- & -- & 0.167 & 0.169 & 0.158 & 0.149 & 0.692 & \textbf{0.364} & 0.873 & 0.916 & 0.79 \\
Custom K/V only & 30\,s & \xmark & \xmark & 0.556 & 0.592 & 0.516 & 0.505 & 0.789 & 0.355 & \textbf{0.890} & \textbf{0.923} & 0.81 \\
+\,DVA & 30\,s & \cmark & \xmark & 0.614 & 0.614 & 0.584 & 0.562 & 0.787 & 0.347 & 0.878 & 0.921 & 0.62 \\
\oc+\,DVA\,+\,ACG (Ours) & \oc30\,s & \oc\cmark & \oc\cmark & \oc\textbf{0.641} & \oc\textbf{0.634} & \oc\textbf{0.623} & \oc\textbf{0.615} & \oc\textbf{0.798} & \oc0.348 & \oc0.889 & \oc\textbf{0.923} & \oc0.87 \\
\midrule
Custom K/V only & 2\,min & \xmark & \xmark & 0.518 & 0.586 & 0.466 & 0.447 & 0.780 & \textbf{0.354} & 0.859 & 0.911 & 0.84 \\
\oc+\,DVA\,+\,ACG (Ours) & \oc2\,min & \oc\cmark & \oc\cmark & \oc\textbf{0.620} & \oc\textbf{0.632} & \oc\textbf{0.583} & \oc\textbf{0.571} & \oc\textbf{0.793} & \oc0.342 & \oc\textbf{0.866} & \oc\textbf{0.914} & \oc0.86 \\
\bottomrule
\end{tabular}}
\caption{\textbf{Results on Deep Forcing}, $n=100$. Columns and bold follow Tab.~\ref{tab:main}, and rows follow Tab.~\ref{tab:comp}.}
\label{tab:deep}
\end{table}

\subsection{Results per Subject}
\label{app:subject}
Tab.~\ref{tab:subject} breaks the $30$\,s results down by subject. \method{} raises the last window over Custom K/V only for every subject, by $0.142$ on average (subject-level bootstrap $95\%$ confidence interval $0.093$ to $0.189$) and in $88$ of the $100$ prompts. The gain is largest for the animals, which show the most identity drift under Custom K/V only, and smallest for the backpack and the RC car, which show little. The Custom anchor column is the DINO-I of the custom anchor image itself, $0.679$ on average. Custom K/V only and \method{} both start from this image and differ in how much of its identity survives to the last window.

\begin{table}[h]
\centering
\small
\begin{tabular}{lcccc}
\toprule
& & \multicolumn{2}{c}{DINO-I, last window} & \\
\cmidrule(lr){3-4}
Subject & Custom anchor & Custom K/V only & \method{} & Gain \\
\midrule
backpack & 0.669 & 0.647 & 0.662 & $+0.015$ \\
cat & 0.695 & 0.425 & 0.636 & $+0.211$ \\
cat2 & 0.788 & 0.481 & 0.702 & $+0.221$ \\
dog & 0.656 & 0.509 & 0.617 & $+0.108$ \\
dog3 & 0.460 & 0.055 & 0.316 & $+0.261$ \\
dog7 & 0.706 & 0.452 & 0.604 & $+0.151$ \\
dog8 & 0.678 & 0.407 & 0.589 & $+0.182$ \\
monster\_toy & 0.611 & 0.494 & 0.554 & $+0.060$ \\
rc\_car & 0.715 & 0.687 & 0.723 & $+0.036$ \\
robot\_toy & 0.810 & 0.589 & 0.761 & $+0.172$ \\
\midrule
All & 0.679 & 0.475 & 0.616 & $+0.142$ \\
\bottomrule
\end{tabular}
\caption{\textbf{Results per subject}, $30$\,s, $10$ prompts each.}
\label{tab:subject}
\end{table}

\subsection{Drift Sensor}
\label{app:sensor}
DVA senses drift with DINOv2-B, the same encoder behind DINO-I. Tab.~\ref{tab:sensor} swaps the sensor for ConvNeXt-B~\citep{convnext} and scores identity with both encoders. The scores change by at most $0.003$.

\begin{table}[h]
\centering
\small
\begin{tabular}{lcc}
\toprule
& \multicolumn{2}{c}{Scored with} \\
\cmidrule(lr){2-3}
Drift sensor & DINO-I & ConvNeXt-I \\
\midrule
\oc DINOv2-B (Ours) & \oc0.635 & \oc0.626 \\
ConvNeXt-B & 0.634 & 0.623 \\
\bottomrule
\end{tabular}
\caption{\textbf{Drift sensor and evaluator}, $30$\,s, $n=100$.}
\label{tab:sensor}
\end{table}

\subsection{Identity over Time}
\label{app:time}
\begin{figure}[h]
\centering
\includegraphics[width=\linewidth]{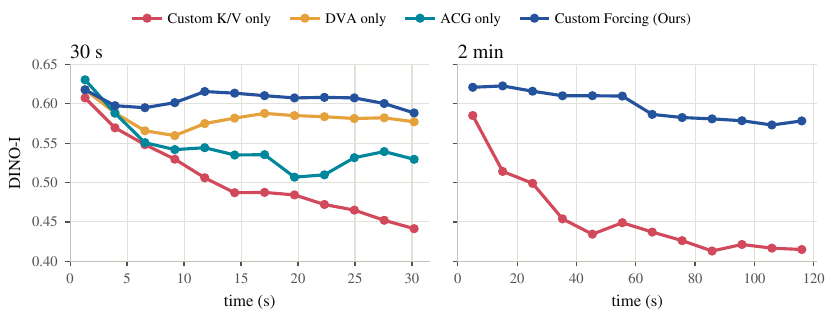}
\caption{\textbf{DINO-I over time}, $n=100$. Left: $30$\,s with each component. Right: $2$\,min.}
\label{fig:time}
\end{figure}

Fig.~\ref{fig:time} plots DINO-I in $12$ equal bins over the video. With Custom K/V only, it declines steadily, from $0.61$ to $0.44$ over $30$\,s and from $0.59$ to $0.41$ over $2$\,min. DVA stops the decline after about $10$\,s, as its amplification responds to the drift, and ACG alone starts highest but still declines. \method{} stays near $0.60$ for $30$\,s and ends the two-minute video at $0.58$.

\subsection{Class Prior}
\label{app:prior}
Section~\ref{sec:nag} observes that the subject of Custom K/V only drifts toward generic instances of its category. To measure this, we compute $s_{\text{prior}}$, which is DINO-I with the frames of the vanilla AR video for the same prompt in place of the reference images, and report it on the same thirds as DINO-I (Tab.~\ref{tab:acg_prior}). With Custom K/V only, $s_{\text{prior}}$ rises from $0.395$ to $0.418$ as DINO-I falls from $0.579$ to $0.475$. Adding ACG lowers $s_{\text{prior}}$ in every third, by $0.015$ to $0.040$, and keeps it near $0.38$, while DINO-I rises by $0.016$ to $0.076$.

\begin{table}[h]
\centering
\small
\begin{tabular}{lcccccc}
\toprule
& \multicolumn{3}{c}{DINO-I $\uparrow$} & \multicolumn{3}{c}{$s_{\text{prior}}$ $\downarrow$} \\
\cmidrule(lr){2-4}\cmidrule(l){5-7}
Method & first & middle & last & first & middle & last \\
\midrule
Custom K/V only & 0.579 & 0.506 & 0.475 & 0.395 & 0.414 & 0.418 \\
\,+\,ACG & \textbf{0.595} & \textbf{0.551} & \textbf{0.550} & \textbf{0.380} & \textbf{0.385} & \textbf{0.378} \\
\bottomrule
\end{tabular}
\caption{\textbf{Similarity to the reference images and to the class prior}, $30$\,s, $n=100$.}
\label{tab:acg_prior}
\end{table}

\subsection{User Study and VLM Judges}
\label{app:user}

\begin{figure}[h]
\centering
\includegraphics[width=\linewidth]{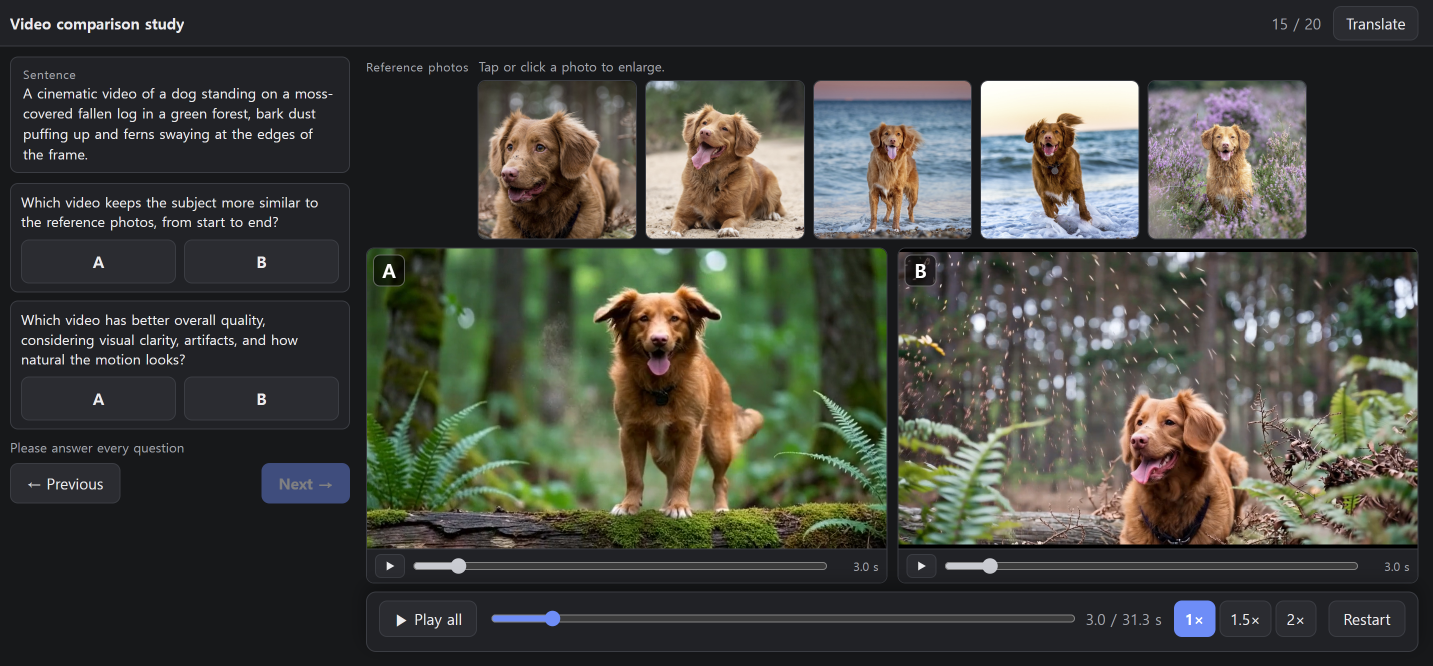}
\caption{\textbf{User study interface.} Participants see the prompt, the reference images  and choose A or B for each question before moving to the next pair.}
\label{fig:user_ui}
\end{figure}

\paragraph{User study.}
We select $20$ of the $100$ prompts, two per subject, and pair the $30$\,s video of \method{} for each prompt with that of each of five baselines, Custom K/V only and the four image- and reference-to-video models, which gives $100$ pairs. Thirty participants took part. Each participant rates $20$ pairs, assigned by a Latin square so that every prompt appears once and every baseline four times, which gives $600$ comparisons and $120$ per baseline. Each screen shows the reference images, the prompt, and the two videos playing in sync, in random order and without method names, and participants can pause, seek, and speed up the videos. For each pair, they answer two questions, ``Which video keeps the subject more similar to the reference photos, from start to end?'' and ``Which video has better overall quality, considering visual clarity, artifacts, and how natural the motion looks?''  Fig.~\ref{fig:user_ui} shows one screen, on which participants compare the two videos and answer both questions  by choosing A or B.

\paragraph{VLM judge.}
We ask Qwen3-VL-235B-A22B-Instruct~\citep{qwen3vl}, Claude Sonnet 5, and Claude Opus 5.5 the same two questions on the same $100$ pairs. Each pair is given as one image that contains the reference images, the prompt, and ten frames of each video at $t = 3, 6, \ldots, 30$\,s in two rows, with \method{} in the top row for half of the pairs of each baseline. The model answers Top or Bottom for each question.

\paragraph{Results.}
Tab.~\ref{tab:user} gives the percentage of judgments that prefer \method{}. For the subject, participants prefer \method{} over Custom K/V only, FramePack, FramePack-F1, and SkyReels-V2 DF ($60.0$ to $64.2\%$) and in $58.5\%$ of all $600$ votes. Against SkyReels-V3, whose 14B backbone reaches a similar whole-video DINO-I, they show no significant preference ($45.8\%$). For quality, they prefer \method{} over FramePack ($70.0\%$) and overall ($56.8\%$), and show no significant preference against Custom K/V only ($45.8\%$), so DVA and ACG improve the identity without a measurable loss in quality.

Qwen3-VL, Claude Sonnet 5, and Claude Opus 5.5 also prefer \method{} overall, in $66$, $63$, and $76\%$ of the pairs for the subject and in $67$, $63$, and $75\%$ for quality. All three prefer it over FramePack-F1 and SkyReels-V2 DF in $65$ to $95\%$ of the pairs. All three VLM judges differ from the participants on SkyReels-V3, which they prefer less than \method{} ($60$ to $90\%$), and they rate the quality of FramePack closer to \method{} than the participants do ($45$ to $65\%$ against $70.0\%$). They disagree with each other on Custom K/V only, which Claude Sonnet 5 prefers for both questions ($40$ and $35\%$), while Qwen3-VL and Claude Opus 5.5 prefer \method{} for the subject ($60$ and $80\%$). Unlike the participants, the VLM judges see ten frames rather than the videos.

\begin{table}[h]
\centering
\small
\resizebox{\linewidth}{!}{%
\setlength{\tabcolsep}{4pt}
\begin{tabular}{@{}l cc cc cc cc@{}}
\toprule
& \multicolumn{2}{c}{Human} & \multicolumn{2}{c}{Qwen3-VL-235B} & \multicolumn{2}{c}{Claude Sonnet 5} & \multicolumn{2}{c}{Claude Opus 5.5} \\
\cmidrule(lr){2-3}\cmidrule(lr){4-5}\cmidrule(lr){6-7}\cmidrule(l){8-9}
\method{} vs. & Subject & Quality & Subject & Quality & Subject & Quality & Subject & Quality \\
\midrule
Custom K/V only & 60.0 & 45.8 & 60.0 & 50.0 & 40.0 & 35.0 & 80.0 & 60.0 \\
FramePack       & 60.0 & 70.0 & 50.0 & 45.0 & 45.0 & 50.0 & 55.0 & 65.0 \\
FramePack-F1    & 62.5 & 57.5 & 65.0 & 70.0 & 80.0 & 75.0 & 95.0 & 85.0 \\
SkyReels-V2 DF  & 64.2 & 59.2 & 75.0 & 80.0 & 80.0 & 75.0 & 90.0 & 85.0 \\
SkyReels-V3     & 45.8 & 51.7 & 80.0 & 90.0 & 70.0 & 80.0 & 60.0 & 80.0 \\
\midrule
All             & 58.5 & 56.8 & 66.0 & 67.0 & 63.0 & 63.0 & 76.0 & 75.0 \\
\bottomrule
\end{tabular}}
\caption{\textbf{User study and VLM judges.} Percentage of judgments that prefer \method{} over each method; $50\%$ means no preference. Human: $120$ votes per cell from $30$ participants. VLM judges: Qwen3-VL-235B, Claude Sonnet 5, and Claude Opus 5.5, each with one judgment on each of $20$ pairs per cell and $100$ in \emph{All}.}
\label{tab:user}
\end{table}

\section{Qualitative Observations}
\label{app:qual}

\paragraph{Two-minute generation.}
In Fig.~\ref{fig:long}, the eyes of the toy under Custom K/V only grow from $24$\,s and rise on stalks by $48$\,s while its body swells, and the gray merle dog turns black tricolor by $48$\,s. \method{} keeps the appearance of both subjects for the full two minutes.

\paragraph{Image- and reference-to-video models.}
In Fig.~\ref{fig:i2v}, FramePack and SkyReels-V3 keep the fur of both animals until $30$\,s as \method{} does, but with ten times its parameters. The image-to-video models receive only the custom anchor, and SkyReels-V3 sees the reference images only in its first $5$\,s. \method{} instead keeps them in its KV cache throughout generation, and its DINO-I stays between $0.616$ and $0.626$ from the first window to the last.

\paragraph{Adding the components.}
In Fig.~\ref{fig:qual}, Custom K/V only raises the ears of the dog and turns the toy into a four-wheeled car by $21$\,s. Adding DVA, which corrects the strength of the reference conditioning, delays the ears to $30$\,s and keeps the toy a robot, but with a different face and shoulders. Adding ACG, which corrects its direction, keeps both subjects as in the reference images.

\section{Additional Ablations}
\label{app:ablations}

\subsection{Injection and Guidance Strength}
\label{app:alphak}
\label{app:alpha}
\label{app:acg}
Tab.~\ref{tab:alphak} varies the injection budget $\alpha_{\text{inj}}$ at $\kappa = 4$. A smaller budget lowers identity ($-0.009$ DINO-I at $180$). A larger one does not raise it, and Subject Consistency, Background Consistency, and Dynamic Degree fall ($0.41 \to 0.33$ at $420$). Identity also drops toward the end of the video. At $420$, DINO-I on the last third of the video is $0.041$ below the first third, against $0.003$ at $300$. Tab.~\ref{tab:acg} varies the guidance strength $\kappa$ at $\alpha_{\text{inj}} = 300$. DINO-I rises up to $\kappa = 4$ and saturates at $5$, where Dynamic Degree already falls to $0.36$. At $\kappa = 6$, CLIP-I, Subject Consistency, and Dynamic Degree fall ($0.41 \to 0.32$), and DINO-I drops toward the end of the video as it does at a large $\alpha_{\text{inj}}$. On both axes, the operating point thus gives the best trade-off between identity and motion.

The blend weight needs no separate sweep. By Eq.~\eqref{eq:acgshare}, when the clip does not bind, $\alpha_{\text{acg}}$ and $\kappa$ act only through $\alpha_{\text{acg}}(\kappa - 1)$, which the $\kappa$ sweep already covers.

\begin{table}[h]
\centering
\footnotesize
\renewcommand{\arraystretch}{0.92}
\setlength{\tabcolsep}{3pt}
\begin{minipage}[t]{0.49\linewidth}
\centering
\resizebox{\linewidth}{!}{%
\begin{tabular}{lccccc}
\toprule
$\alpha_{\text{inj}}$ & DINO-I & CLIP-I & Subj.\ Cons. & Bg.\ Cons. & Dynamic \\
\midrule
$180$ & 0.626 & 0.801 & \textbf{0.930} & \textbf{0.947} & 0.41 \\
$240$ & 0.632 & \textbf{0.803} & 0.929 & 0.946 & 0.40 \\
$300^{\ast}$ & \textbf{0.635} & \textbf{0.803} & 0.928 & 0.945 & 0.41 \\
$360$ & 0.632 & \textbf{0.803} & 0.923 & 0.942 & 0.35 \\
$420$ & 0.623 & 0.801 & 0.918 & 0.941 & 0.33 \\
\bottomrule
\end{tabular}}
\caption{\textbf{Injection budget $\alpha_{\text{inj}}$} at $\kappa = 4$, $n=100$.}
\label{tab:alphak}
\end{minipage}\hfill
\begin{minipage}[t]{0.49\linewidth}
\centering
\resizebox{\linewidth}{!}{%
\begin{tabular}{lccccc}
\toprule
$\kappa$ & DINO-I & CLIP-I & Subj.\ Cons. & Bg.\ Cons. & Dynamic \\
\midrule
$2$ & 0.623 & \textbf{0.805} & 0.924 & \textbf{0.945} & 0.39 \\
$3$ & 0.629 & 0.804 & 0.926 & \textbf{0.945} & 0.40 \\
$4^{\ast}$ & \underline{0.635} & 0.803 & \textbf{0.928} & \textbf{0.945} & 0.41 \\
$5$ & \textbf{0.636} & 0.802 & 0.926 & 0.944 & 0.36 \\
$6$ & 0.626 & 0.797 & 0.920 & 0.943 & 0.32 \\
\bottomrule
\end{tabular}}
\caption{\textbf{Guidance strength $\kappa$} at $\alpha_{\text{inj}} = 300$, $n=100$.}
\label{tab:acg}
\end{minipage}
\par\smallskip
{\footnotesize $^{\ast}$\,The operating point of \method{}. Bold marks the best value in each column, and an underline marks the operating point where it ranks second. Dynamic Degree is not bolded.}
\end{table}

\subsection{Composition of the Custom K/V}
\label{app:rladder}
\label{app:rladder9}
\begin{figure}[h]
\centering
\includegraphics[width=\linewidth]{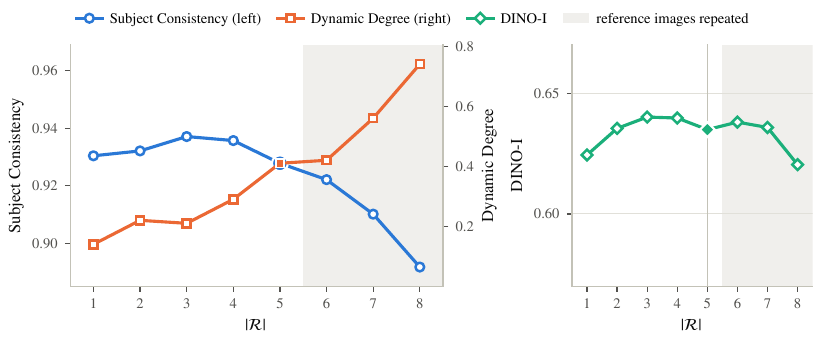}
\caption{\textbf{Number of reference anchors at $\Nslot = 9$}, $30$\,s, $n=100$. Reference images are repeated in the shaded region. Left: Subject Consistency and Dynamic Degree, with the two axes aligned at the default $|\Rset| = 5$. Right: DINO-I.}
\label{fig:rladder}
\end{figure}

Fig.~\ref{fig:rladder} varies $|\Rset|$ at fixed $\Nslot = 9$, so every reference image added replaces a copy of the custom anchor. DINO-I barely moves. It stays between $0.620$ and $0.640$ and is lowest at the two ends, $|\Rset| = 1$ and $|\Rset| = 8$. What $|\Rset|$ changes is the balance between the two kinds of anchor. With few reference images, the copies of the custom anchor dominate and the video barely moves, with a Dynamic Degree of $0.14$ at $|\Rset| = 1$. With many, the one remaining copy cannot hold the scene. At $|\Rset| = 8$, Subject and Background Consistency fall below those of Custom K/V only ($0.892$ and $0.933$ against $0.913$ and $0.939$). The higher Dynamic Degree there ($0.74$) reflects the scene drifting rather than the subject moving. $|\Rset| = 5$ is the smallest value at which Dynamic Degree reaches that of Custom K/V only ($0.41$), and the largest that uses each reference image once. Both consistency scores remain above those of Custom K/V only at this value.

\subsection{Generated Images in Place of the Reference Images}
\label{app:s9}
Tab.~\ref{tab:fill} replaces the reference images in the custom K/V with images that FLUX.2\,[klein] generates from the same reference images and prompt with other seeds, without selection. G5K4 fills the five reference slots with seeds $1$ to $5$ and keeps four copies of the custom anchor. G8K1 uses seeds $1$ to $8$ and keeps one copy. No reference image enters the cache, although the generated images carry the subject from them, and DINO-I is still measured against the reference images. With G5K4, \method{} scores only $0.005$ DINO-I below the default R5K4, but Dynamic Degree drops from $0.41$ to $0.30$. With G8K1, DINO-I is $0.020$ lower. In every setting, DVA and ACG raise DINO-I most in the last window.

\begin{table}[h]
\centering
\small
\renewcommand{\arraystretch}{1.0}
\begin{tabular}{llccccc}
\toprule
& & \multicolumn{4}{c}{DINO-I} & \\
\cmidrule(lr){3-6}
& & all & first & middle & last & Dynamic \\
\midrule
\multirow{2}{*}{R5K4} & Custom K/V only & 0.541 & 0.579 & 0.506 & 0.475 & 0.41 \\
& \oc\method{} (Ours) & \oc\textbf{0.635} & \oc0.619 & \oc0.626 & \oc\textbf{0.616} & \oc0.41 \\
\midrule
\multirow{2}{*}{G5K4} & Custom K/V only & 0.571 & 0.601 & 0.550 & 0.523 & 0.34 \\
& \oc\method{} (Ours) & \oc0.630 & \oc\textbf{0.628} & \oc\textbf{0.628} & \oc0.606 & \oc0.30 \\
\midrule
\multirow{2}{*}{G8K1} & Custom K/V only & 0.556 & 0.577 & 0.538 & 0.512 & 0.40 \\
& \oc\method{} (Ours) & \oc0.615 & \oc0.611 & \oc0.595 & \oc0.583 & \oc0.56 \\
\bottomrule
\end{tabular}
\caption{\textbf{What the custom K/V holds}, $30$\,s, $n=100$. R, G, and K count reference images, generated images, and copies of the custom anchor. Bold marks the best value in each column, and Dynamic Degree is not bolded.}
\label{tab:fill}
\end{table}

\subsection{Constant Amplification and Subject Mask}
\label{app:const}
\label{sec:exp:const}
\begin{figure}[h]
\centering
\includegraphics[width=\linewidth]{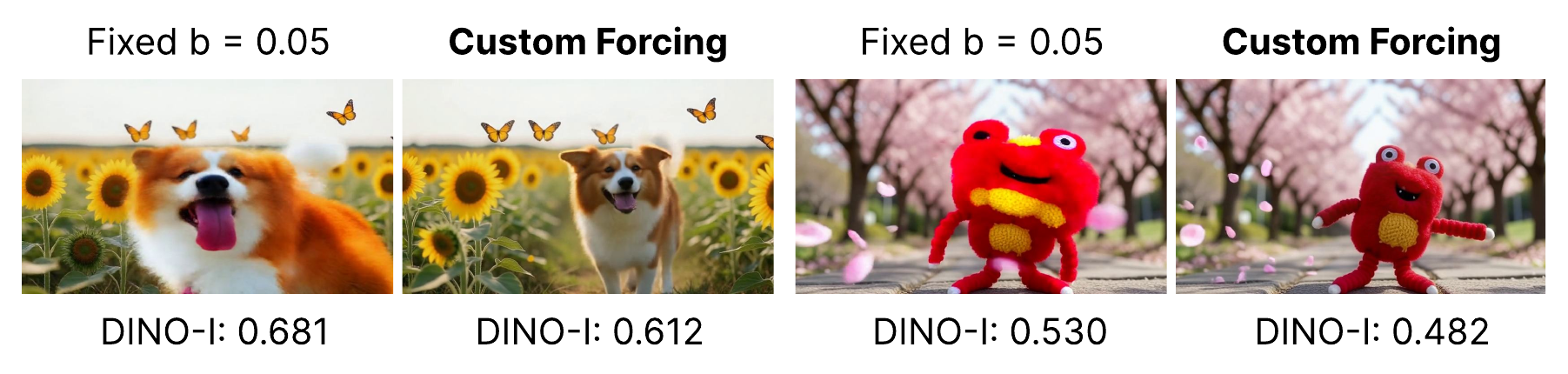}
\caption{\textbf{A constant amplification raises DINO-I but breaks the scene.} Frames at $17.5$\,s of two prompts, whose DINO-I rises from $0.612$ to $0.681$ and from $0.482$ to $0.530$.}
\label{fig:constgain}
\end{figure}

Tab.~\ref{tab:const} replaces the measured drift $\bmag(t)$ with a constant and removes the subject mask. A constant $\bmag \equiv 0.05$ raises DINO-I by $0.016$, but the subject grows until it fills the frame (Fig.~\ref{fig:constgain}). CLIP-T, Subject Consistency, and Dynamic Degree all drop. Larger constants break the video. The adaptive amplification avoids this because it stays small. Measured $\bmag(t)$ averages $0.022$, which gives a target amplification of about $1.3$ against $3$ for $\bmag \equiv 0.05$, and exceeds $0.05$ in only $14\%$ of chunks. Removing the subject mask lowers DINO-I, Subject Consistency, and Dynamic Degree. Background tokens carry no identity, and amplifying them ties the video to the backgrounds of the reference images.

\begin{table}[h]
\centering
\small
\renewcommand{\arraystretch}{1.0}
\begin{tabular}{lcccccc}
\toprule
& DINO-I & CLIP-I & CLIP-T & Subj.\ Cons. & Bg.\ Cons. & Dynamic \\
\midrule
\oc\method{} (Ours) & \oc0.635 & \oc0.803 & \oc\textbf{0.353} & \oc\textbf{0.928} & \oc\textbf{0.945} & \oc0.41 \\
$\bmag \equiv 0.05$ & \textbf{0.651} & \textbf{0.804} & 0.335 & 0.887 & 0.926 & 0.28 \\
$\bmag \equiv 0.1$ & 0.403 & 0.729 & 0.329 & 0.793 & 0.875 & 0.08 \\
$\bmag \equiv 0.2$ & 0.158 & 0.643 & 0.309 & 0.691 & 0.823 & 0.02 \\
no subject mask & 0.622 & 0.803 & 0.352 & 0.913 & 0.936 & 0.30 \\
\bottomrule
\end{tabular}
\caption{\textbf{Adaptive amplification and subject mask}, $30$\,s, $n=100$. The largest constant, $0.2$, equals the cap $\bcap$. Bold marks the best value in each column, and Dynamic Degree is not bolded.}
\label{tab:const}
\end{table}

\section{Limitations}
\label{app:limit}
\method{} controls identity for the subject as a whole, not for its parts. The drift signal $s(t)$ compares the whole subject with the reference images, and DVA sets the amplification of every subject token of the reference anchors from this single signal. A change confined to one small part, such as a wheel of the RC car or a strap of the backpack, may therefore change $s(t)$ little, and \method{} cannot target that part on its own. The fidelity of \method{} is also bounded by the frozen backbone. \method{} changes only how the backbone attends to the anchors, so appearance that the backbone cannot render remains out of reach. Finally, we evaluate one subject per video for up to two minutes. Multiple subjects and longer horizons remain for future work.

\section{Evaluation Prompts}
\label{app:prompts}
Table~\ref{tab:prompts} lists all evaluation prompts. Every prompt names the subject only by its category. Three pairs of subjects share the same ten prompts (cat and cat2, dog and dog3, dog7 and dog8), so the $100$ prompts contain $70$ distinct sentences. The ten prompts of robot\_toy describe five scenes twice, once with the toy moving and once with it holding still.

{\small
\setlength{\LTpre}{4pt}\setlength{\LTpost}{0pt}\setlength{\LTcapwidth}{\linewidth}
\begin{longtable}{@{}>{\raggedright\arraybackslash}p{0.25\linewidth}@{\hspace{0.02\linewidth}}p{0.73\linewidth}@{}}
\caption{\textbf{Evaluation prompts}, grouped by subject. The noun used in the prompts is in parentheses.}\label{tab:prompts}\\
\toprule
Slug & Prompt \\
\midrule
\endfirsthead
\toprule
Slug & Prompt (continued) \\
\midrule
\endhead
\bottomrule
\endfoot
\multicolumn{2}{@{}l}{\cellcolor{black!8}\textbf{backpack}\hspace{0.5em}(``backpack'')} \\*
\texttt{\footnotesize 00\_park\_bench} & A cinematic video of a backpack resting on a weathered wooden park bench in an autumn park, orange leaves turning overhead and drifting past behind it. \\*
\texttt{\footnotesize 01\_forest\_rock} & A cinematic video of a backpack sitting on a flat mossy rock beside a winding forest trail, dappled sunlight flickering through the canopy and ferns swaying at the edges. \\*
\texttt{\footnotesize 02\_city\_sidewalk} & A cinematic video of a backpack standing upright on a quiet city sidewalk lined with shopfronts, traffic drifting past far behind it and shop awnings rippling in the breeze. \\*
\texttt{\footnotesize 03\_beach} & A cinematic video of a backpack resting on flat open sand at a wide beach, fine sand streaming past it and waves breaking far behind. \\*
\texttt{\footnotesize 04\_mountain\_ridge} & A cinematic video of a backpack sitting on a rocky mountain ridge above a deep valley, wind tugging at its straps and clouds drifting across the valley below. \\*
\texttt{\footnotesize 05\_wooden\_floor} & A cinematic video of a backpack standing upright on a polished wooden floor in a sunlit room, light shifting across the boards as clouds pass the window. \\*
\texttt{\footnotesize 06\_cherry\_blossom} & A cinematic video of a backpack resting on a stone path under cherry trees in full bloom, pink petals drifting down around it and branches swaying overhead. \\*
\texttt{\footnotesize 07\_library\_desk} & A cinematic video of a backpack resting on a long wooden desk in a bright library, dust motes floating in the window light and a page lifting in the draft. \\*
\texttt{\footnotesize 08\_cobblestone} & A cinematic video of a backpack standing upright on a cobblestone street in an old town, light shifting across the stones as clouds pass overhead. \\*
\texttt{\footnotesize 09\_train\_platform} & A cinematic video of a backpack standing upright on an empty train platform under a long canopy, a train passing far down the line and litter stirring in the draft. \\
\addlinespace[3pt]
\multicolumn{2}{@{}l}{\cellcolor{black!8}\textbf{cat}\hspace{0.5em}(``cat'')} \\*
\texttt{\footnotesize 00\_windowsill} & A cinematic video of a cat stretching on a wooden windowsill in a bright room, its back arching and a curtain stirring beside it. \\*
\texttt{\footnotesize 01\_garden\_path} & A cinematic video of a cat trotting along a stone garden path between low flower beds, its tail swaying and petals trembling in the breeze. \\*
\texttt{\footnotesize 02\_rooftop} & A cinematic video of a cat walking across a flat sunlit rooftop above a quiet city, its fur stirring in the wind and clouds drifting behind it. \\*
\texttt{\footnotesize 03\_wooden\_desk} & A cinematic video of a cat sitting on a wooden desk beside a stack of books in a study, a page lifting in the draft as it lands. \\*
\texttt{\footnotesize 04\_cherry\_blossom} & A cinematic video of a cat running down a cherry blossom street in full spring bloom, pink petals scattering in its wake. \\*
\texttt{\footnotesize 05\_sofa} & A cinematic video of a cat lying on a cozy sofa in a warmly lit living room, cushions dipping beneath it and dust motes floating in the light. \\*
\texttt{\footnotesize 06\_courtyard} & A cinematic video of a cat trotting across a sunlit cobblestone courtyard in an old town, light shifting across the stones as clouds pass. \\*
\texttt{\footnotesize 07\_park\_lawn} & A cinematic video of a cat running across a wide grassy park lawn with trees behind it, grass parting around its paws and leaves turning overhead. \\*
\texttt{\footnotesize 08\_stone\_wall} & A cinematic video of a cat walking along a low stone wall beside a country lane, its tail counterbalancing each step and wildflowers nodding below. \\*
\texttt{\footnotesize 09\_terracotta\_patio} & A cinematic video of a cat sitting on a sunlit terracotta patio beside potted plants, potted leaves stirring as it passes. \\
\addlinespace[3pt]
\multicolumn{2}{@{}l}{\cellcolor{black!8}\textbf{cat2}\hspace{0.5em}(``cat'')} \\*
\texttt{\footnotesize 00\_windowsill} & A cinematic video of a cat stretching on a wooden windowsill in a bright room, its back arching and a curtain stirring beside it. \\*
\texttt{\footnotesize 01\_garden\_path} & A cinematic video of a cat trotting along a stone garden path between low flower beds, its tail swaying and petals trembling in the breeze. \\*
\texttt{\footnotesize 02\_rooftop} & A cinematic video of a cat walking across a flat sunlit rooftop above a quiet city, its fur stirring in the wind and clouds drifting behind it. \\*
\texttt{\footnotesize 03\_wooden\_desk} & A cinematic video of a cat sitting on a wooden desk beside a stack of books in a study, a page lifting in the draft as it lands. \\*
\texttt{\footnotesize 04\_cherry\_blossom} & A cinematic video of a cat running down a cherry blossom street in full spring bloom, pink petals scattering in its wake. \\*
\texttt{\footnotesize 05\_sofa} & A cinematic video of a cat lying on a cozy sofa in a warmly lit living room, cushions dipping beneath it and dust motes floating in the light. \\*
\texttt{\footnotesize 06\_courtyard} & A cinematic video of a cat trotting across a sunlit cobblestone courtyard in an old town, light shifting across the stones as clouds pass. \\*
\texttt{\footnotesize 07\_park\_lawn} & A cinematic video of a cat running across a wide grassy park lawn with trees behind it, grass parting around its paws and leaves turning overhead. \\*
\texttt{\footnotesize 08\_stone\_wall} & A cinematic video of a cat walking along a low stone wall beside a country lane, its tail counterbalancing each step and wildflowers nodding below. \\*
\texttt{\footnotesize 09\_terracotta\_patio} & A cinematic video of a cat sitting on a sunlit terracotta patio beside potted plants, potted leaves stirring as it passes. \\
\addlinespace[3pt]
\multicolumn{2}{@{}l}{\cellcolor{black!8}\textbf{dog}\hspace{0.5em}(``dog'')} \\*
\texttt{\footnotesize 00\_beach} & A cinematic video of a dog running along the shoreline of a sandy beach at sunset, paws kicking up sand, waves rolling in and out, fur blowing in the sea breeze. \\*
\texttt{\footnotesize 01\_forest} & A cinematic video of a dog trotting down a winding forest trail, dappled sunlight flickering through swaying leaves, dust and pollen drifting in the air. \\*
\texttt{\footnotesize 02\_city\_street} & A cinematic video of a dog walking briskly along a busy city sidewalk, cars passing by and pedestrians blurring in the background, neon reflections on wet pavement. \\*
\texttt{\footnotesize 03\_sunflower\_field} & A cinematic video of a dog bounding playfully through a vast field of blooming sunflowers, petals and leaves swaying in the wind, butterflies fluttering up. \\*
\texttt{\footnotesize 04\_mountain\_top} & A cinematic video of a dog standing on a rocky mountain peak, fur ruffled by strong gusts of wind, clouds drifting across the valley below. \\*
\texttt{\footnotesize 05\_snowy\_clearing} & A cinematic video of a dog trotting through a snowy clearing ringed by dark evergreen pines, snow puffing up from its paws, flakes drifting down from the branches. \\*
\texttt{\footnotesize 06\_wooden\_bridge} & A cinematic video of a dog trotting across an old wooden bridge, a river rushing and sparkling below, planks creaking, mist rising from the water. \\*
\texttt{\footnotesize 07\_eiffel\_tower} & A cinematic video of a dog walking in front of the Eiffel Tower, tail wagging, tourists strolling past and pigeons taking flight, soft golden hour light. \\*
\texttt{\footnotesize 08\_sofa} & A cinematic video of a dog lying on a cozy sofa in a warmly lit living room, curtains gently swaying by an open window, dust motes floating in the light. \\*
\texttt{\footnotesize 09\_cherry\_blossom} & A cinematic video of a dog walking down a cherry blossom street in spring, pink petals falling and drifting on the breeze, branches swaying overhead. \\
\addlinespace[3pt]
\multicolumn{2}{@{}l}{\cellcolor{black!8}\textbf{dog3}\hspace{0.5em}(``dog'')} \\*
\texttt{\footnotesize 00\_beach} & A cinematic video of a dog running along the shoreline of a sandy beach at sunset, paws kicking up sand, waves rolling in and out, fur blowing in the sea breeze. \\*
\texttt{\footnotesize 01\_forest} & A cinematic video of a dog trotting down a winding forest trail, dappled sunlight flickering through swaying leaves, dust and pollen drifting in the air. \\*
\texttt{\footnotesize 02\_city\_street} & A cinematic video of a dog walking briskly along a busy city sidewalk, cars passing by and pedestrians blurring in the background, neon reflections on wet pavement. \\*
\texttt{\footnotesize 03\_sunflower\_field} & A cinematic video of a dog bounding playfully through a vast field of blooming sunflowers, petals and leaves swaying in the wind, butterflies fluttering up. \\*
\texttt{\footnotesize 04\_mountain\_top} & A cinematic video of a dog standing on a rocky mountain peak, fur ruffled by strong gusts of wind, clouds drifting across the valley below. \\*
\texttt{\footnotesize 05\_snowy\_clearing} & A cinematic video of a dog trotting through a snowy clearing ringed by dark evergreen pines, snow puffing up from its paws, flakes drifting down from the branches. \\*
\texttt{\footnotesize 06\_wooden\_bridge} & A cinematic video of a dog trotting across an old wooden bridge, a river rushing and sparkling below, planks creaking, mist rising from the water. \\*
\texttt{\footnotesize 07\_eiffel\_tower} & A cinematic video of a dog walking in front of the Eiffel Tower, tail wagging, tourists strolling past and pigeons taking flight, soft golden hour light. \\*
\texttt{\footnotesize 08\_sofa} & A cinematic video of a dog lying on a cozy sofa in a warmly lit living room, curtains gently swaying by an open window, dust motes floating in the light. \\*
\texttt{\footnotesize 09\_cherry\_blossom} & A cinematic video of a dog walking down a cherry blossom street in spring, pink petals falling and drifting on the breeze, branches swaying overhead. \\
\addlinespace[3pt]
\multicolumn{2}{@{}l}{\cellcolor{black!8}\textbf{dog7}\hspace{0.5em}(``dog'')} \\*
\texttt{\footnotesize 00\_beach} & A cinematic video of a dog running along a wide sandy beach, sand kicking up from its paws and the sea breeze lifting its fur. \\*
\texttt{\footnotesize 01\_forest\_trail} & A cinematic video of a dog trotting down a winding forest trail, sunlight flickering through the canopy and pollen drifting in the air. \\*
\texttt{\footnotesize 02\_city\_sidewalk} & A cinematic video of a dog walking along a quiet city sidewalk lined with shopfronts, traffic moving slowly far behind it. \\*
\texttt{\footnotesize 03\_wheat\_field} & A cinematic video of a dog running through a golden wheat field under an open sky, wheat stalks parting and bending in the wind. \\*
\texttt{\footnotesize 04\_mountain\_ridge} & A cinematic video of a dog standing on a rocky mountain ridge above a deep valley, wind ruffling its coat and clouds drifting across the valley behind it. \\*
\texttt{\footnotesize 05\_fallen\_log} & A cinematic video of a dog standing on a moss-covered fallen log in a green forest, bark dust puffing up and ferns swaying at the edges of the frame. \\*
\texttt{\footnotesize 06\_autumn\_park} & A cinematic video of a dog running through a park path buried in orange autumn leaves, leaves scattering behind it and branches swaying overhead. \\*
\texttt{\footnotesize 07\_sofa} & A cinematic video of a dog lying on a cozy sofa in a warmly lit living room, curtains swaying gently by an open window and dust motes floating in the light. \\*
\texttt{\footnotesize 08\_cherry\_blossom} & A cinematic video of a dog walking down a cherry blossom street in full spring bloom, pink petals drifting down and branches swaying overhead. \\*
\texttt{\footnotesize 09\_desert\_dune} & A cinematic video of a dog walking across a wind-sculpted sandy desert dune under a wide open sky, fine sand streaming off the crest behind it. \\
\addlinespace[3pt]
\multicolumn{2}{@{}l}{\cellcolor{black!8}\textbf{dog8}\hspace{0.5em}(``dog'')} \\*
\texttt{\footnotesize 00\_beach} & A cinematic video of a dog running along a wide sandy beach, sand kicking up from its paws and the sea breeze lifting its fur. \\*
\texttt{\footnotesize 01\_forest\_trail} & A cinematic video of a dog trotting down a winding forest trail, sunlight flickering through the canopy and pollen drifting in the air. \\*
\texttt{\footnotesize 02\_city\_sidewalk} & A cinematic video of a dog walking along a quiet city sidewalk lined with shopfronts, traffic moving slowly far behind it. \\*
\texttt{\footnotesize 03\_wheat\_field} & A cinematic video of a dog running through a golden wheat field under an open sky, wheat stalks parting and bending in the wind. \\*
\texttt{\footnotesize 04\_mountain\_ridge} & A cinematic video of a dog standing on a rocky mountain ridge above a deep valley, wind ruffling its coat and clouds drifting across the valley behind it. \\*
\texttt{\footnotesize 05\_fallen\_log} & A cinematic video of a dog standing on a moss-covered fallen log in a green forest, bark dust puffing up and ferns swaying at the edges of the frame. \\*
\texttt{\footnotesize 06\_autumn\_park} & A cinematic video of a dog running through a park path buried in orange autumn leaves, leaves scattering behind it and branches swaying overhead. \\*
\texttt{\footnotesize 07\_sofa} & A cinematic video of a dog lying on a cozy sofa in a warmly lit living room, curtains swaying gently by an open window and dust motes floating in the light. \\*
\texttt{\footnotesize 08\_cherry\_blossom} & A cinematic video of a dog walking down a cherry blossom street in full spring bloom, pink petals drifting down and branches swaying overhead. \\*
\texttt{\footnotesize 09\_desert\_dune} & A cinematic video of a dog walking across a wind-sculpted sandy desert dune under a wide open sky, fine sand streaming off the crest behind it. \\
\addlinespace[3pt]
\multicolumn{2}{@{}l}{\cellcolor{black!8}\textbf{monster\_toy}\hspace{0.5em}(``toy'')} \\*
\texttt{\footnotesize 00\_wooden\_floor} & A cinematic video of a toy walking across a polished wooden floor in a sunlit room, its limbs swinging with each step and light shifting across the boards. \\*
\texttt{\footnotesize 01\_grass\_lawn} & A cinematic video of a toy running across a short grass lawn in a sunny garden, its arms swinging and grass rippling around its feet. \\*
\texttt{\footnotesize 02\_park\_path} & A cinematic video of a toy marching along a paved park path lined with autumn trees, leaves turning overhead and drifting past behind it. \\*
\texttt{\footnotesize 03\_stone\_steps} & A cinematic video of a toy sitting on worn stone steps on a quiet old street, its arms reaching for the next step. \\*
\texttt{\footnotesize 04\_beach} & A cinematic video of a toy standing on flat open sand at a wide beach, sand puffing up beneath it and the sea breeze stirring its fur. \\*
\texttt{\footnotesize 05\_bed} & A cinematic video of a toy sitting on a neatly made bed in a bright bedroom, the duvet dipping and springing back beneath it. \\*
\texttt{\footnotesize 06\_cherry\_blossom} & A cinematic video of a toy standing on a stone path under cherry trees in full bloom, its arm swinging in a wave and pink petals drifting down around it. \\*
\texttt{\footnotesize 07\_picnic\_blanket} & A cinematic video of a toy sitting on a checkered picnic blanket in a green meadow, its arms swinging side to side and grass rippling around it. \\*
\texttt{\footnotesize 08\_garden\_path} & A cinematic video of a toy walking along a stone garden path between low flower beds, its feet lifting in turn and petals trembling in the breeze. \\*
\texttt{\footnotesize 09\_woven\_rug} & A cinematic video of a toy sitting on a patterned woven rug in a warmly lit room, it rolling over once and warm light shifting across the rug. \\
\addlinespace[3pt]
\multicolumn{2}{@{}l}{\cellcolor{black!8}\textbf{rc\_car}\hspace{0.5em}(``toy'')} \\*
\texttt{\footnotesize 00\_garage\_floor} & A cinematic video of a toy rolling across a smooth concrete floor in a bright garage, its wheels spinning and reflections sliding across the polished floor, camera tracking low alongside. \\*
\texttt{\footnotesize 01\_city\_sidewalk} & A cinematic video of a toy driving along a quiet city sidewalk lined with shopfronts, its wheels turning and shopfronts sliding past behind it, camera panning to keep pace. \\*
\texttt{\footnotesize 02\_dirt\_track} & A cinematic video of a toy driving along a flat dirt track across an open field, loose grit spraying from its wheels, camera tracking smoothly alongside. \\*
\texttt{\footnotesize 03\_beach} & A cinematic video of a toy driving across flat firm sand at a wide beach, sand streaming out behind its wheels, camera gliding low alongside. \\*
\texttt{\footnotesize 04\_cobblestone} & A cinematic video of a toy driving along a cobblestone street in an old town, the body bouncing over the stones, camera dollying alongside. \\*
\texttt{\footnotesize 05\_wooden\_floor} & A cinematic video of a toy rolling across a polished wooden floor in a sunlit room, its reflection sliding along the boards, camera tracking low from the side. \\*
\texttt{\footnotesize 06\_parking\_lot} & A cinematic video of a toy driving across a wide empty asphalt parking lot, painted lines sweeping past beneath it, camera craning up as it passes. \\*
\texttt{\footnotesize 07\_park\_path} & A cinematic video of a toy driving along a paved park path lined with trees, dappled light sweeping across its body, camera tracking low alongside. \\*
\texttt{\footnotesize 08\_stone\_plaza} & A cinematic video of a toy driving across a wide stone plaza in a historic square, pigeons scattering far ahead of it, camera slowly orbiting as it drives. \\*
\texttt{\footnotesize 09\_race\_track} & A cinematic video of a toy driving along a painted asphalt race track with curbing along its edge, curb stripes flashing past beneath it, camera tracking low alongside. \\
\addlinespace[3pt]
\multicolumn{2}{@{}l}{\cellcolor{black!8}\textbf{robot\_toy}\hspace{0.5em}(``toy'')} \\*
\texttt{\footnotesize 00\_wooden\_floor\allowbreak\_\_motion} & A cinematic video of a toy rolling across a polished wooden floor in a sunlit room, light shifting across the boards as clouds pass the window. \\*
\texttt{\footnotesize 01\_wooden\_floor\allowbreak\_\_static} & A cinematic video of a toy standing on a polished wooden floor in a sunlit room, light shifting across the boards as clouds pass the window. \\*
\texttt{\footnotesize 02\_park\_path\allowbreak\_\_motion} & A cinematic video of a toy rolling across a paved park path lined with autumn trees, leaves turning overhead and dappled light sweeping over it. \\*
\texttt{\footnotesize 03\_park\_path\allowbreak\_\_static} & A cinematic video of a toy standing on a paved park path lined with autumn trees, leaves turning overhead and dappled light sweeping over it. \\*
\texttt{\footnotesize 04\_stone\_plaza\allowbreak\_\_motion} & A cinematic video of a toy rolling across a wide stone plaza in a historic square, light shifting across the stone as clouds pass overhead. \\*
\texttt{\footnotesize 05\_stone\_plaza\allowbreak\_\_static} & A cinematic video of a toy standing on a wide stone plaza in a historic square, light shifting across the stone as clouds pass overhead. \\*
\texttt{\footnotesize 06\_grass\_lawn\allowbreak\_\_motion} & A cinematic video of a toy rolling across a short grass lawn in a sunny garden, blades swaying around it and light shifting as clouds pass. \\*
\texttt{\footnotesize 07\_grass\_lawn\allowbreak\_\_static} & A cinematic video of a toy standing on a short grass lawn in a sunny garden, blades swaying around it and light shifting as clouds pass. \\*
\texttt{\footnotesize 08\_race\_track\allowbreak\_\_motion} & A cinematic video of a toy rolling across a painted asphalt race track with curbing along its edge, heat shimmering off the asphalt and banners rippling beyond the curbing. \\*
\texttt{\footnotesize 09\_race\_track\allowbreak\_\_static} & A cinematic video of a toy standing on a painted asphalt race track with curbing along its edge, heat shimmering off the asphalt and banners rippling beyond the curbing. \\
\end{longtable}
}

\end{document}

%% file: math_commands.tex
\usepackage{amsmath,amsfonts,bm}

\def\eqref#1{equation~\ref{#1}}

\def\1{\bm{1}}

\DeclareMathAlphabet{\mathsfit}{\encodingdefault}{\sfdefault}{m}{sl}
\SetMathAlphabet{\mathsfit}{bold}{\encodingdefault}{\sfdefault}{bx}{n}

